\documentclass[a4paper,fleqn]{cas-dc}

\usepackage[numbers]{natbib}

\usepackage{tikz}
\usepackage{pgfplots}
\pgfplotsset{compat=1.18}
\usepackage{subcaption}
\usetikzlibrary{fit, backgrounds}
\usetikzlibrary{positioning,calc,arrows.meta,shapes.geometric}

\usepackage[table]{xcolor}
\definecolor{lightgrayrow}{RGB}{245,245,245}
\usepackage{amsmath}

\usepackage[section]{placeins}
\usepackage{flafter}
\usepackage{tabularx}
\usepackage{chngcntr}

\def\tsc#1{\csdef{#1}{\textsc{\lowercase{#1}}\xspace}}
\tsc{WGM}
\tsc{QE}
\tsc{EP}
\tsc{PMS}
\tsc{BEC}
\tsc{DE}

\begin{document}
\let\WriteBookmarks\relax

\shorttitle{} 
\shortauthors{Bashir, A. et~al.}

\title [mode = title]{Sim-to-Real Reinforcement Learning for Path-Safe Target Reaching in Robotic Harvesting of Hydroponic Lettuce}                      

\author[1,4]{Al Bashir}[orcid=0000-0003-2168-2840]
\ead{albashir@tamu.edu}
\credit{Conceptualization, Methodology, Formal analysis, Data curation, Validation, Writing original – review \& editing, draft, Visualization}

\author[1,4]{Partho Ghose}[orcid=0000-0002-5840-2008]
\ead{partho.ghose@tamu.edu}
\credit{Writing – review \& editing}

\author[1,4]{Yibin Wang}[orcid=0000-0002-8761-5387]
\ead{yibinwang@tamu.edu}
\credit{Writing – review \& editing}

\author[1]{Robert G. Hardin}
\ead{robert.hardin@ag.tamu.edu}
\credit{Writing – review \& editing}

\author[2]{Joshua Peeples}[orcid=0000-0001-7861-7789]
\ead{jpeeples@tamu.edu}
\credit{Writing – review \& editing}

\author[3, 4]{Genhua Niu}
\ead{genhua.niu@ag.tamu.edu}
\credit{Writing – review \& editing}

\author[1,4]{Azlan Zahid}[orcid=0000-0001-6202-8680]
\cormark[1]
\ead{azlan.zahid@tamu.edu}
\credit{Supervision, Funding acquisition, Writing – review \& editing}

\affiliation[1]{organization={Department of Biological \& Agricultural Engineering, Texas A\&M University},
                city={College Station},
                state={TX},
                postcode={77843}, 
                country={USA}}

\affiliation[2]{organization={Department of Electrical \& Computer Engineering, Texas A\&M University},
                city={College Station},
                state={TX},
                postcode={77843}, 
                country={USA}}           

\affiliation[3]{organization={Department of Horticultural Sciences, Texas A\&M University},
                city={College Station},
                state={TX},
                postcode={77843}, 
                country={USA}}
                
\affiliation[4]{organization={Texas A\&M AgriLife Research, Texas A\&M University System},
                city={Dallas},
                state={TX},
                postcode={75252}, 
                country={USA}}
                
                
\cortext[cor1]{Corresponding author}

\begin{abstract}
Robotic harvesting requires precise manipulation to ensure accurate positioning of the end-effector, especially in a constraint workspace. This study introduces the Path-Safe Approach Proximal Policy Optimization (PSA-PPO), a target-conditioned local reaching framework designed for robotic harvesting of nutrient film technique (NFT)-hydroponic lettuce in greenhouses. The proposed PSA-PPO preserves the standard clipped PPO objective while introducing a harvesting-oriented policy formulation that integrates target-conditioned observations, corridor-aware spatial constraints, trajectory-history-based safety encoding, and task-specific success criteria for constrained robotic manipulation. PSA-PPO was trained entirely in Isaac Lab and successfully transferred to a physical UR10e robotic arm without any real-world policy fine-tuning. In a controlled physical benchmark against 150 scheduled targets, PSA-PPO achieved the highest reaching rate of 94.7\%, outperforming Cartesian+TOTG (86.0\%), RRTConnect (88.0\%), and RRT* (80.0\%), while reducing median RMS jerk by 39.8–73.0\% relative to the classical planners during the final approach segment. A formulation ablation evaluated over 200 simulation targets demonstrates that path-safety and side-corridor shaping are the principal drivers of harvesting-compatible performance: removing either term substantially degrades strict success even when terminal endpoint accuracy is preserved or improved, confirming that optimization toward the endpoint alone is an insufficient objective for constrained crop-proximal manipulation. Finally, PSA-PPO was integrated with an RGB-D vision module and an end-effector and evaluated as a complete greenhouse harvesting system across 144 lettuce harvesting attempts, achieving 94.4\% vision-guided target reaching and 86.1\% complete harvesting success.
These results position PSA-PPO as an effective simulation-trained harvesting motion controller that balances target-reaching reliability and low dynamic excitation. 
\end{abstract}


\begin{keywords}
Robotic harvesting \sep Reinforcement learning \sep Sim-to-real \sep
Path-safe control \sep Hydroponic lettuce \sep Motion planning \sep Isaac Lab
\end{keywords}
\maketitle

\section{Introduction}

Reliable manipulator motion remains a critical challenge in robotic harvesting because crop localization, target approach, and physical crop interaction must be executed within highly constrained and variable agricultural environments~\cite{Rajendran2024, Liu2024FieldRobots}. Unlike structured industrial manipulation, crop targets vary continuously in geometry, accessibility, and are surrounded by deformable plant material and cultivation infrastructure~\cite{Jin2024RoboticArms, Zhao2026HarvestPlanning,Hernandez2025Reconfigurable}. These constraints become especially important in controlled-environment agriculture, where dense crop layouts restrict manipulator workspace while increasing the requirement for repeatable autonomous operation~\cite{VANHENTEN2013170}.
Hydroponic lettuce presents a unique challenge in terms of manipulation. The harvesting tool must approach the stem--root region, just above the cultivation infrastructure, establish the right cutting angle, sever the plant, retain the detached lettuce, and transfer it without disturbing neighboring crops or rigid cultivation equipment~\cite{YU2026104362}. Early research on lettuce harvesting demonstrated the effectiveness of machine-vision-guided manipulation within structured production systems~\cite{Cho2002Lettuce}. More recent studies have continued to improve crop-specific harvesting mechanisms and the interactions of end effectors, highlighting the importance of combining crop localization with subsequent robotic actions~\cite{Yu2026LettuceEndEffector, bashir2025development}. For instance, the field-tested Vegebot integrated learned perception, force-feedback control, and a specialized end effector designed for iceberg lettuce harvesting~\cite{Birrell2020Lettuce}.  However, these studies do not address how to convert a vision-derived cutting point for lettuce into a repeatable motion that complies with the geometry and clearance constraints of a complex agricultural environment.

\begin{figure*}
\centering
\includegraphics[width=.95\textwidth]{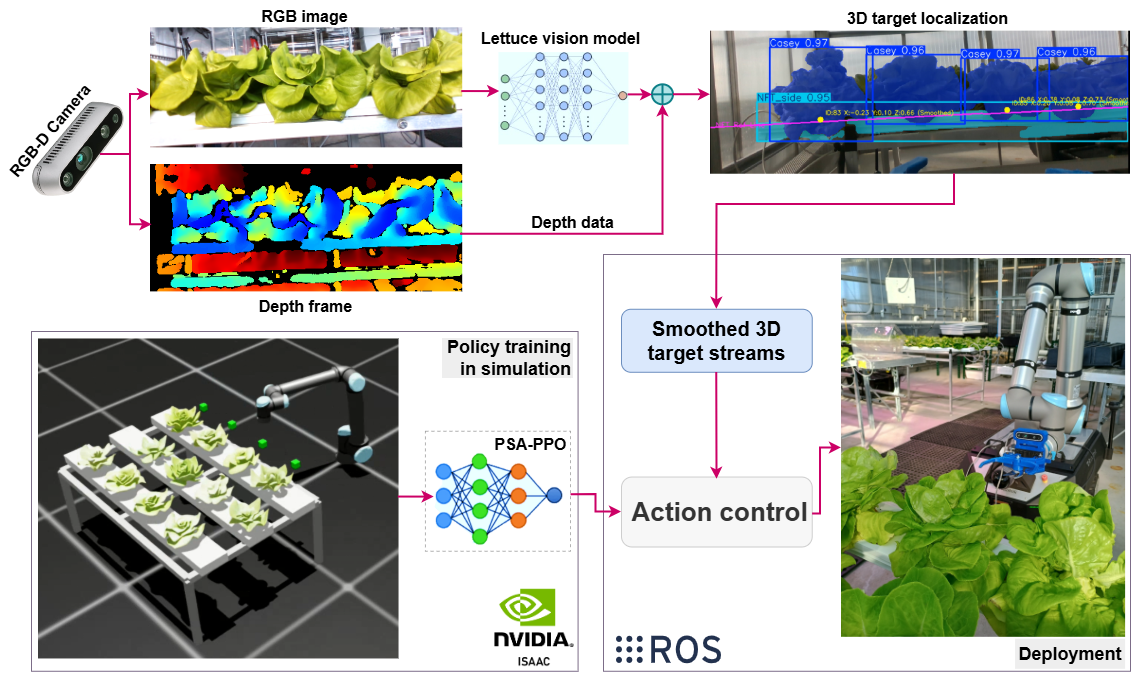}
\caption{Overview of the proposed sim-to-real PSA-PPO harvesting framework.
The robot-mounted RGB-D camera provides RGB imagery and aligned depth to the
lettuce perception module, which generates temporally smoothed 3D cutting-point targets. PSA-PPO is trained entirely in Isaac Lab for target-conditioned, path-safe side-approach reaching. During physical deployment, the frozen policy and vision-derived targets are integrated through the ROS action-control layer to command the UR10e for side-approach harvesting of nutrient film technique (NFT)-grown lettuce.}
\label{fig:overview}
\end{figure*}

Manipulator motion in agricultural harvesting is still predominantly generated using inverse kinematics, sampling-based motion planning, trajectory optimization, or visual servoing.
Researchers have adapted these methods to crop-specific manipulation requirements. For example, Ma et al.~\cite{Ma2024BIT} developed an improved BIT*-based planner~\cite{Karaman2011RRTStar} that jointly considered path quality, execution efficiency, and dynamic impact. Rong et al.~\cite{Rong2025TomatoPlanning} coordinated manipulator and lifting-joint motion to improve tomato-harvesting reachability, while Kang et al.~\cite{Kang2025BroccoliPlanning} proposed a BIT*-based Time-Optimal and Path-Optimal (TOPO) planner that combines improved sampling and edge checking with path optimization to reduce planning time and generate shorter trajectories for selective broccoli harvesting.  Recent studies have similarly modified Informed-RRT*~\cite{Karaman2011RRTStar} for smooth winter-jujube harvesting~\cite{Huang2025JujubePlanning} and developed a sampling-based planner for high-quality motion generation in unstructured tea canopies~\cite{Huang2025TeaPlanning}. At the system level, visual-servoing approaches~\cite{Chaumette2006VisualServo} remain common for cucumber harvesting ~\cite{Park2023Cucumber,Huang2026CucumberHVSC}. Classical planners~\cite{Sucan2012OMPL, Fresnillo2023MoveIt, Kunz2012TOTG} therefore provides strong geometric trajectory and can achieve high terminal precision, but agricultural manipulation requires simultaneous consideration of reachability, obstacle exposure, trajectory smoothness, perception uncertainty, execution robustness, and crop-specific approach geometry ~\cite{He2025HandEye,Zhao2026HarvestPlanning}. Additionally, these planners are vulnerable to singularities, sensitive to noise which often produce unstable motion and inefficient trajectories~\cite{levine2018learning, rajendran2024towards}.

Learning-based policy control offers a complementary approach for achieving smooth motion through learning loops between the robot agent and its environment. Proximal Policy Optimization (PPO)~\cite{Schulman2017PPO} has become a widely used method in Deep Reinforcement Learning (DRL) for continuous robotic control~\cite {Tang2025DRLRobotics}. It uses a policy-gradient method that constrains policy updates through a clipped surrogate objective. However, this mechanism does not explicitly enforce geometric or trajectory-safety constraints; constraint-aware policy optimization remains a distinct problem~\cite{Achiam2017CPO}. For harvesting, a policy rewarded primarily for terminal target acquisition may reach the cutting point without preserving an acceptable approach direction, tool orientation, or intermediate clearance. Addressing this limitation requires a task formulation that represents the relevant spatial relationships, evaluates trajectory-level violations, and distinguishes terminal reaching from harvesting-compatible motion rather than relying on the standard PPO objective alone. A further significant challenge in DRL training is acquiring the necessary interaction experience of robot through trial and error. Policy optimization involves repeated robot motions and resets, substantial execution time, and potentially unsafe exploratory actions. These challenges are particularly critical in agricultural environments, where exploratory actions can damage delicate biological materials, harvesting infrastructure, or the manipulator itself~\cite{DulacArnold2021RealWorldRL, Tang2025DRLRobotics}. These limitations highlight the need to develop manipulation DRL policies without relying on exploratory policy optimization in a physical robot.

Simulation-based policy learning offers a fundamentally different development path. Instead of gathering the extensive interaction data required for policy optimization on a physical harvesting robot, both exploration and policy learning can be carried out in a virtual physics simulator, allowing the learned policy to be evaluated on hardware later. GPU-native simulation has made this approach increasingly feasible. For instance, NVIDIA Isaac Lab~\cite{Mittal2025IsaacLab}, a unified GPU-accelerated framework for reinforcement learning, showcases massively parallel robot-policy interactions directly on the GPU. 


However, simulation-trained policies must overcome the reality gap yielded by imperfect dynamics, actuator behavior, sensing, and calibration variation. Previous research has addressed these discrepancies through visual and dynamics randomization ~\cite{Tobin2017DomainRandomization, Peng2018DynamicsRandomization} and physical parameter identification ~\cite{Sobanbabu2025SysID}. For example, Subedi et al.~\cite{Subedi2026FindFruit} demonstrated simulation-only RL transfer for occlusion-aware plant manipulation by separating learned behavior from lower-level compliant control.  Kim et al.~\cite{Kim2025PepperIL} used visuomotor imitation learning for autonomous pepper harvesting in unstructured fields, while Rong et al.~\cite{Rong2026ActionPolicy} combined position-based visual alignment with a corrective action policy for greenhouse tomato harvesting. However, there remains limited evidence for policies that simultaneously incorporate crop-specific approach geometry, trajectory-history clearance, simulation-only policy optimization, controlled physical comparison with multiple classical planners, and end-to-end harvesting in complex agricultural setups. Moreover, success in simulation alone does not guarantee physical reliability; evaluations of paired robots and policies have highlighted the importance of addressing control and observation discrepancies ~\cite{Li2025SIMPLER}. For effective crop-proximal harvesting, it is essential to preserve the learned control interface while also accommodating the constraints of physical calibration and execution. This research gap is crucial to greenhouse-grown lettuce because successful reaching cannot be defined solely by the final tool center point (TCP) error. The cutter must approach the stem region from a suitable horizontal direction, remain compatible with the target-height geometry, preserve the required tool orientation, limit wrist--TCP descent near the cultivation structure, and avoid undesirably abrupt local motion. Hence, this control problem is consequently one of harvesting-compatible local constrained reaching, rather than generic point-to-point motion alone.

To address these gaps, this study develops \emph{Path-Safe Approach PPO (PSA-PPO)}, a target-conditioned local reaching formulation for hydroponic lettuce harvesting. Rather than modifying PPO's optimization rule, PSA-PPO reformulates the learning task through geometry-conditioned safe initialization, target- and clearance-aware observations, coupled reaching and motion-safety rewards, and trajectory-history-dependent strict capture and termination criteria for collision-avoidance harvesting-specific trajectories. The policy is trained entirely in Isaac Lab and transferred without real-world policy fine-tuning, using calibrated target transformation, a frozen Cartesian deployment correction, and an independent ROS execution and safety interface. The study then examines whether harvesting-compatible behavior learned in simulation is retained on a physical UR10e, how it compares with
classical motion planners, and how it performs within a complete vision-guided greenhouse harvesting system.

The main contributions of this study are:
\begin{enumerate}
    \item A target-conditioned path-safe local reaching formulation, named PSA-PPO, is developed for lettuce harvesting. It incorporates spatial constraints, trajectory-level safety awareness, and harvesting-specific task objectives.
 
    \item The PSA-PPO is trained entirely in simulation, without any physical rollouts used for policy optimization or fine-tuning. The trained policy is evaluated against the identical 200 targets within the simulation, highlighting contributions such as path-safety shaping, side-corridor shaping, and motion regularization.
    
    \item The simulation-trained policy is transferred to a physical UR10e robotic arm and is benchmarked against Cartesian+TOTG, RRTConnect, and RRT* algorithms over 150 common targets. This comparison assesses task reliability, terminal precision, target convergence, physical motion dynamics, joint tracking, and trajectory geometry, revealing the trade-offs involved in the controller.
    
    \item PSA-PPO is integrated with an RGB-D-based lettuce vision module for target streaming and a end-effector mechanism in a complete greenhouse harvesting pipeline. Across 144 actual lettuce harvesting attempts, the system achieved 94.4\% target reaching success.
\end{enumerate}

\section{Methodology}
\label{sec:methodology}

The proposed PSA-PPO motion controller framework integrates with a previously developed lettuce cutting-point localization perception module and cut-and-hold mechanism \cite{11219194,bashir2025development}, creating a complete pipeline. The policy is entirely trained in a simulation environment and then transferred to the physical robot without any fine-tuning required in the real world. Figure \ref{fig:overview} illustrates the resulting perception-learning-deployment pipeline.

\subsection{3D Target Streaming}
\label{subsec:vision_module}
The perception interface extended our previously developed lettuce segmentation and occluded cutting-point localization method \cite{11219194} to RGB-D target streaming. The vision method provided the image-space cutting-point pixel $(u_c, v_c)$. Using depth aligned to the RGB frame, the corresponding depth $Z_c$ and RGB camera intrinsics $(f_x, f_y, c_x, c_y)$ were used to back-project the cutting point into the camera coordinate frame. A fixed camera-frame $Y$-axis offset $d_y$ was then applied to provide clearance from the NFT channel:
\begin{equation}
\mathbf{p}_{cut}^{cam}
=
\begin{bmatrix}
\dfrac{(u_c-c_x)Z_c}{f_x}\\[5pt]
\dfrac{(v_c-c_y)Z_c}{f_y}-d_y\\[5pt]
Z_c
\end{bmatrix},
\qquad
d_y=5~\mathrm{mm}
\label{eq:camera_cutting_point}
\end{equation}

To overcome frame-to-frame variation caused by depth noise and segmentation fluctuations, cutting-point estimates were temporally associated and averaged before robot execution. Let $\mathbf{s}_t^{(j)}=[u_t^{(j)},v_t^{(j)}]^\top$ denote the image-plane tracking center of segmented mask $j$ at frame $t$, and let $\hat{\mathbf{s}}_{t-1}^{(k)}$ denote the latest center associated with active track $k$. Each detection was matched to its nearest existing track:
\begin{equation}
k^\ast(j)
=
\arg\min_{k\in\mathcal{K}_{t-1}}
\left\|
\mathbf{s}_t^{(j)}
-
\hat{\mathbf{s}}_{t-1}^{(k)}
\right\|_2 ,
\label{eq:lettuce_track_assignment}
\end{equation}
\noindent where the assignment was accepted only when the image-plane distance was below $\tau_p=50$ pixels; otherwise, a new track was initialized. Each active track maintained a fixed-length buffer of 3D cutting-point estimates, from which the streamed target was computed as:
\begin{equation}
\bar{\mathbf{p}}_{cut}^{(k)}
=
\frac{1}{N_b}
\sum_{n=1}^{N_b}
\mathbf{p}_{cut,n}^{(k)},
\qquad
N_b=15
\label{eq:lettuce_target_smoothing}
\end{equation}
The averaged point was published as a ROS \texttt{PointStamped} message in the camera optical frame. The buffer was cleared after publication, and tracks not observed in the current frame were removed to prevent stale targets from propagating to the harvesting controller.

\subsection{Simulation-Trained Policy for Path-Safe Side-Approach Target Reaching}
\label{subsec:sim_reaching_policy}

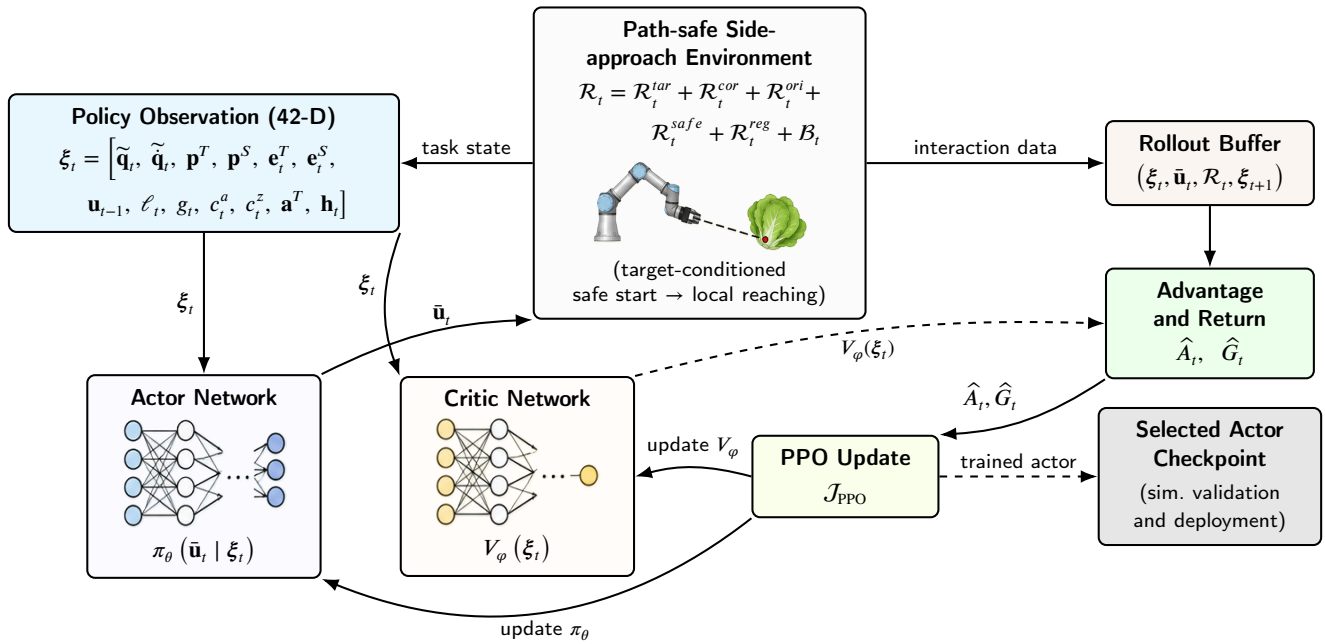
\begin{figure*}
\centering
\resizebox{\textwidth}{!}{%
\begin{tikzpicture}[
    font=\small,
    >=Latex,
    box/.style={
        draw,
        rounded corners=3pt,
        thick,
        align=center,
        inner sep=5pt
    },
    netbox/.style={
        draw,
        rounded corners=3pt,
        thick,
        align=center,
        inner sep=5pt,
        minimum height=1.7cm,
        minimum width=3.2cm
    },
    line/.style={->, thick},
    dline/.style={->, thick, dashed}
]


\node[box, fill=cyan!8, text width=5cm, minimum height=1.0cm] (obs) at (0,3.2)
{
\textbf{Policy Observation (42-D)}\\[2pt]
$\begin{aligned}
\boldsymbol{\xi}_t
=
\left[
\widetilde{\mathbf{q}}_t,\;
\widetilde{\dot{\mathbf{q}}}_t,\;
\mathbf{p}^{T},\;
\mathbf{p}^{S},\;
\mathbf{e}^{T}_t,\;
\mathbf{e}^{S}_t,\; \right.\\
\left. \mathbf{u}_{t-1},\;
\ell_t,\;
g_t,\;
c_t^{a},\;
c_t^{z},\;
\mathbf{a}^{T},\;
\mathbf{h}_t \right]
\end{aligned}$\\[-1pt]
};

\node[box, fill=gray!3, text width=4.2cm, minimum height=3cm] (env) at (6.8,3.2)
{
\textbf{Path-safe Side-approach Environment}\\[3pt]
$\begin{aligned}
\mathcal{R}_t =
\mathcal{R}^{tar}_t+
\mathcal{R}^{cor}_t+
\mathcal{R}^{ori}_t+\\
\mathcal{R}^{safe}_t+
\mathcal{R}^{reg}_t+
\mathcal{B}_t
\end{aligned}$\\[4pt]
\includegraphics[width=3.0cm]{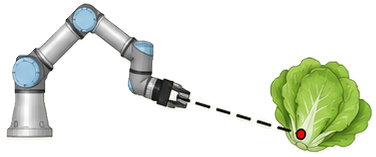}\\
\footnotesize (target-conditioned \\safe start $\rightarrow$ local reaching)\\
};

\node[box, fill=brown!8, text width=2.5cm, minimum height=1cm] (rollout) at (13.8,3.2)
{
\textbf{Rollout Buffer}\\[2pt]
$\left(
\boldsymbol{\xi}_t,\bar{\mathbf{u}}_t,
\mathcal{R}_t,\boldsymbol{\xi}_{t+1}
\right)$
};

\node[box, fill=green!8, text width=2.5cm, minimum height=1.45cm] (adv) at (13.8,1.0)
{
\textbf{Advantage and Return}\\[2pt]
$\widehat{A}_t,\;\widehat{G}_t$
};

\node[box, fill=lime!8, text width=2.2cm, minimum height=1cm] (ppo) at (8.8,-1.1)
{
\textbf{PPO Update}\\[2pt]
$\mathcal{J}_{\mathrm{PPO}}$
};

\node[netbox, fill=blue!2, text width=2.5cm, minimum height=2cm] (actor) at (0,-1.1)
{
\textbf{Actor Network}\\[3pt]
\includegraphics[width=2.3cm]{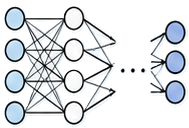}\\
$\pi_{\theta}
\left(
\bar{\mathbf{u}}_t \mid \boldsymbol{\xi}_t
\right)$\\[2pt]
};

\node[netbox, fill=orange!2, text width=2.5cm, minimum height=2cm] (critic) at (4.3,-1.1)
{
\textbf{Critic Network}\\[3pt]
\includegraphics[width=2.3cm]{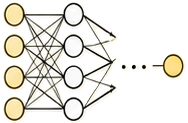}\\
$V_{\varphi}
\left(
\boldsymbol{\xi}_t
\right)$\\[2pt]
};

\node[box, fill=black!10, text width=2.7cm, minimum height=1cm] (deploy) at (13.8,-1.1)
{
\textbf{Selected Actor Checkpoint}\\[2pt]
\footnotesize (sim. validation and deployment)
};


\draw[line] (env.west) -- node[above] {\footnotesize task state} (obs.east);

\draw[line] (obs.south) -- node[left] {$\boldsymbol{\xi}_t$} (actor.north);

\draw[line] (obs.south east) to[bend right=20]
node[midway, above left] {$\boldsymbol{\xi}_t$} (critic.north west);

\draw[line] (actor.north east) to[bend left=12]
node[midway, above right, align=center] {$\bar{\mathbf{u}}_t$}
(env.south west);

\draw[line] (env.east) -- node[above] {\footnotesize interaction data} (rollout.west);

\draw[line] (rollout.south) -- (adv.north);

\draw[line] (adv.south west) to[bend left=15]
node[midway, above left] {$\widehat{A}_t,\widehat{G}_t$} (ppo.north east);

\draw[dline] (critic.north east) to[bend left=8]
node[midway, below] {\footnotesize $V_{\varphi}(\boldsymbol{\xi}_t)$}
(adv.west);


\draw[line] (ppo.west) to[bend right=15]
node[midway, above] {\footnotesize update $V_{\varphi}$}
(critic.east);

\draw[line] (ppo.south west) to[bend left=30]
node[midway, below] {\footnotesize update $\pi_{\theta}$}
(actor.south east);


\draw[dline] (ppo.east) to[bend left=0]
node[midway, above] {\footnotesize trained actor}
(deploy.west);

\end{tikzpicture}%
}
\caption{Training architecture of the path-safe, side-corridor approach policy (PSA-PPO) for the UR10e manipulator. Each target $\mathbf{p}^{T}$ defines a horizontal approach direction $\mathbf{a}^{T}$ and same-height staging reference $\mathbf{p}^{S}$ and determines a geometry-conditioned safe-start configuration. The actor receives the 42-dimensional policy observation $\boldsymbol{\xi}_t$ and produces a six-dimensional action
$\bar{\mathbf{u}}_t$, which is bounded, filtered, and scaled before
application. The environment evaluates target closure, corridor following,
tool orientation, wrist--TCP path safety, and motion regularity. Rollout data and critic estimates are used to compute returns and advantages for the clipped PPO update. The selected actor checkpoint is subsequently
validated in simulation and exported for physical deployment.}
\label{fig:ppo_architecture_lettuce}
\end{figure*}
PSA-PPO was trained entirely in Isaac Lab~\cite{Mittal2025IsaacLab} as a
target-conditioned policy for the final local harvesting approach. Rather
than learning the complete motion from a fixed global home configuration,
the task was decomposed into deterministic target-conditioned
pre-positioning followed by local reaching. Policy optimization was
therefore restricted to the final approach region, where target closure,
side-entry geometry, tool orientation, and wrist--TCP clearance are most
relevant to harvesting. Figure~\ref{fig:ppo_architecture_lettuce} summarizes the training
architecture.

Let $\mathbf{p}^{T}=[p_x^{T},p_y^{T},p_z^{T}]^\top$ denote the target TCP position. The target-conditioned horizontal approach direction and same-height staging reference were defined as:
\begin{equation}
\mathbf{a}^{T}
=
\frac{
[p_x^{T},\,p_y^{T},\,0]^{\top}
}{
\sqrt{(p_x^{T})^2+(p_y^{T})^2}
},
\qquad
\mathbf{p}^{S}
=
\mathbf{p}^{T}-d_a\mathbf{a}^{T}
\label{eq:lettuce_approach_geometry}
\end{equation}
\noindent where $d_a$ is the staging-to-target distance. Because $\mathbf{a}^{T}$ has no vertical component, $\mathbf{p}^{S}$ and $\mathbf{p}^{T}$ have the same height.Each target was sampled before initialization, and its radius, azimuth, and height were used to deterministically select a geometry-conditioned safe-start joint configuration. The robot was therefore initialized near the corresponding local approach region and started the side approach corridor motion towards the target.

The reaching task was implemented as a finite-horizon reinforcement-learning
process. The environment maintained trajectory-level variables required for
reward and termination evaluation, whereas the actor received the following
42-dimensional observation at policy step $t$:
\begin{equation}
\begin{split}
\boldsymbol{\xi}_t
= &
\left[ \widetilde{\mathbf{q}}_t,\; \widetilde{\dot{\mathbf{q}}}_t,\; \mathbf{p}^{T},\; \mathbf{p}^{S},\; \mathbf{e}^{T}_t,\; \mathbf{e}^{S}_t,\;\mathbf{u}_{t-1},\;\right.\\
& \quad \quad \quad \quad \quad \quad \left.  \ell_t,\; g_t,\; c_t^{a},\; c_t^{z},\; \mathbf{a}^{T},\; \mathbf{h}_t \right]
\end{split}
\label{eq:lettuce_policy_obs}
\end{equation}
\noindent where $\widetilde{\mathbf{q}}_t$ and $\widetilde{\dot{\mathbf{q}}}_t$ are the scaled joint positions and velocities, $\mathbf{e}^{T}_t=\mathbf{p}^{T}-\mathbf{p}^{tcp}_t$ and $\mathbf{e}^{S}_t=\mathbf{p}^{S}-\mathbf{p}^{tcp}_t$, and $\mathbf{u}_{t-1}$ is the previous action. The normalized corridor progress was $\ell_t=((\mathbf{p}^{tcp}_t-\mathbf{p}^{S})^\top\mathbf{a}^{T})/d_a$, while $g_t$ indicated activation of the staging-to-target corridor. Tool orientation was represented by $c_t^{a}=\mathbf{f}_t^\top\mathbf{a}^{T}$ and $c_t^{z}=|\mathbf{f}_t^\top\mathbf{e}_z|$, where $\mathbf{f}_t$ is the tool forward axis and $\mathbf{e}_z=[0,0,1]^\top$. Thus, the desired side-approach orientation corresponds to $c_t^{a}\rightarrow1$ and $c_t^{z}\rightarrow0$. The clearance vector $\mathbf{h}_t\in\mathbb{R}^{5}$ contains the three wrist-link clearances, TCP clearance, and their minimum value relative to the target-height safety plane. The resulting observation contained 42 elements.

The stochastic actor produced the normalized raw action
$\bar{\mathbf{u}}_t\in\mathbb{R}^{6}$, which was clipped and temporally
filtered before being converted to an incremental joint-position command:
\begin{equation}
\begin{split}
\mathbf{u}_t = &
(1-\alpha)\mathbf{u}_{t-1}
+
\alpha\,
\operatorname{clip}
\left(
\bar{\mathbf{u}}_t,-1,1
\right),\\
\mathbf{q}^{cmd}_t
&=
\operatorname{clip}_{\mathbf{q}^{min},\mathbf{q}^{max}}
\left(
\mathbf{q}_t+s_u\mathbf{u}_t
\right)
\end{split}
\label{eq:lettuce_action_processing}
\end{equation}
\noindent where $\alpha$ is the temporal-filter coefficient, $s_u$ is the
action scale, and $\mathbf{q}^{min}$ and $\mathbf{q}^{max}$ are the UR10e
soft joint limits.

To separate target closure from corridor deviation, let $\mathbf{a}^{\perp}=[-a_y^{T},a_x^{T},0]^\top$. The target-frame errors were-
\begin{equation}
\begin{split}
& e_t^{ax}
=
(\mathbf{p}^{T}-\mathbf{p}^{tcp}_t)^\top\mathbf{a}^{T},
\qquad
e_t^{lat}
=
(\mathbf{p}^{T}-\mathbf{p}^{tcp}_t)^\top\mathbf{a}^{\perp}, \\ 
& e_t^{z}
=
p_z^{T}-p_{z,t}^{tcp}
\end{split}
\label{eq:lettuce_target_frame_error}
\end{equation}

Path safety was evaluated from the three wrist links and TCP, denoted collectively by $\mathcal{C}$. Their minimum clearance, instantaneous below-plane violation, and maximum accumulated violation were-
\begin{equation}
\begin{split}
&h_t^{min}
=
\min_{i\in\mathcal{C}}
\left(z_t^{(i)}-p_z^{T}\right),
\qquad
b_t
=
\max(0,-h_t^{min}),\\
& b_t^{max}
=
\max_{0\leq\tau\leq t} b_\tau
\end{split}
\label{eq:lettuce_path_safety}
\end{equation}
Thus, a trajectory that temporarily passed below the target-height plane could not be classified as path-safe solely from its final configuration.

The instantaneous reward combined target progress, side-approach corridor tracking, orientation, path safety, motion regularization, and terminal capture:
\begin{equation}
\mathcal{R}_t
=
\mathcal{R}^{tar}_t
+
\mathcal{R}^{cor}_t
+
\mathcal{R}^{ori}_t
+
\mathcal{R}^{safe}_t
+
\mathcal{R}^{reg}_t
+
\mathcal{B}_t
\label{eq:lettuce_reward}
\end{equation}
The target component rewarded TCP-to-target closure, axial progress, forward corridor progression, and entry into increasingly tight target
neighborhoods. The corridor component penalized lateral and vertical departure, axial overshoot, and loitering near the staging region without
continued target closure. The orientation component promoted alignment of the tool axis with $\mathbf{a}^{T}$ while penalizing vertical tilt. The
safety component penalized instantaneous and accumulated target-height-plane violation, and the regularization component penalized action magnitude,
action-rate variation, action acceleration, and joint velocity.

A valid capture required simultaneous positional, orientation, and instantaneous-clearance conditions, whereas strict capture additionally required the complete learned path and initial safe-start configuration to satisfy the path-safety tolerance:
\begin{equation}
\begin{split}
&\chi_t =
\mathbb{I}
\Big[
|e_t^{ax}|<\varepsilon_{ax}
\land
|e_t^{lat}|<\varepsilon_{lat}
\land
|e_t^{z}|<\varepsilon_z\\
&
\quad \quad\land\;
d_t^{T}<\varepsilon_d
\land
c_t^{a}>\varepsilon_a
\land
c_t^{z}<\varepsilon_v
\land
h_t^{min}>0
\Big],\\[1pt]
&\chi_t^{strict} =
\chi_t\,
\mathbb{I}
\left[
b_0\leq\varepsilon_{safe}
\land
b_t^{max}\leq\varepsilon_{safe}
\right]
\end{split}
\label{eq:lettuce_capture_conditions}
\end{equation}
\noindent where $d_t^{T}=\|\mathbf{p}^{T}-\mathbf{p}^{tcp}_t\|_2$ and $b_0$ is the below-plane violation at initialization. The capture reward was $\mathcal{B}_t=c_{b1}\chi_t+c_{b2}\chi_t^{strict}$, with the larger contribution assigned to path-safe capture. Successful termination required $\chi_t^{strict}=1$ for $N_s$ consecutive policy steps, whereas $b_t^{max}>\varepsilon_{hard}$ terminated the episode early.

The actor and critic were trained using PPO with the RSL-RL backend in Isaac Lab. Let $\pi_{\theta}(\bar{\mathbf{u}}_t|\boldsymbol{\xi}_t)$ and $V_{\varphi}(\boldsymbol{\xi}_t)$ denote the actor and critic, respectively. Generalized advantage estimation provided $\widehat{A}_t$ and $\widehat{G}_t$, and the actor was updated using:
\begin{equation}
\begin{split}
\mathcal{J}_{\mathrm{PPO}}(\theta) = &
\mathbb{E}_{t}
\left[
\min
\left(
\rho_t(\theta)\widehat{A}_t,\right.\right.\\
& \quad \quad \quad
 \left.\left. \operatorname{clip}
\left(
\rho_t(\theta), 1-\epsilon, 1+\epsilon
\right)
\widehat{A}_t
\right)
\right]
\end{split}
\label{eq:lettuce_ppo_objective}
\end{equation}

where $\rho_t(\theta)
=
\frac{
\pi_{\theta}(\bar{\mathbf{u}}_t|\boldsymbol{\xi}_t)}{ \pi_{\theta_{\mathrm{old}}}(\bar{\mathbf{u}}_t|\boldsymbol{\xi}_t)}$. The critic was optimized by value-function regression, with entropy regularization retained during training.

Training targets were sampled over the practical front workspace using horizontal polar coordinates and area-uniform radial sampling. A hard-region mixture increased exposure to far-radius, negative-lateral, low- and high-height, and near-boundary azimuth configurations. Candidate checkpoints were subsequently evaluated on an independent random-workspace simulation set; however  crop cutting and plant interaction were not used for policy optimization. Workspace bounds, safe-start settings, component definitions, activation gates, reward coefficients, capture and safety thresholds, network parameters and the terms removed in each formulation ablation are reported in Appendix~\ref{app:A} and \ref{app:B}.
\subsection{End-Effector Integration}
\label{subsec:end_effector}
\begin{figure*}
\centering
\includegraphics[width=.95\textwidth]{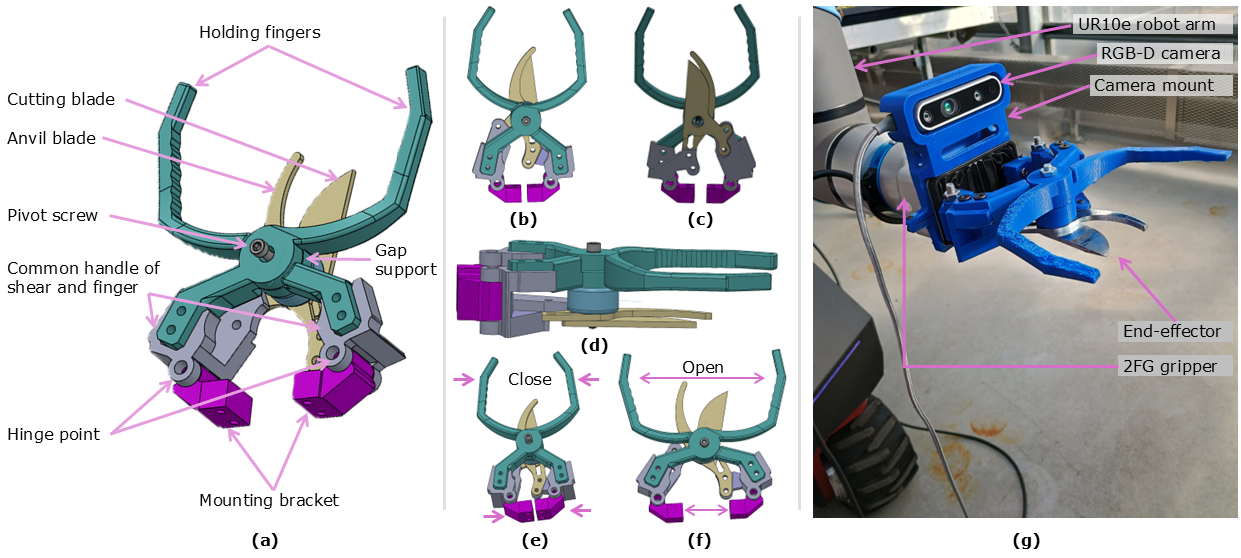}
\caption{Redesigned cut-and-hold lettuce harvesting end effector. (a--f) SolidWorks components and assembled geometry, including top, bottom, side, open, and closed configurations; (g) fabricated mechanism mounted on the OnRobot 2FG7 and UR10e manipulator.} 
\label{fig:end-effector}
\end{figure*}
The harvesting tool adapted our previously developed lettuce cut-and-hold end effector~\cite{bashir2025development} for direct actuation by an
OnRobot 2FG7 gripper. The redesigned components were modeled in SolidWorks 2026 (Dassault Systèmes, Vélizy-Villacoublay, France) and 3D-printed using fused deposition modeling (FDM) with polylactic acid (PLA) filament (Figure~\ref{fig:end-effector}). The original 2FG7 fingers were replaced by custom brackets carrying paired holding fingers and shear blades while retaining the first-class lever cutting principle. A single gripper closing action therefore actuated the shear mechanism and simultaneously retained the detached lettuce for transfer.

\subsection{ROS-Based Deployment and Harvest Sequencing}
\label{subsec:ros_harvesting_control}

\begin{figure*}
\centering
\resizebox{\textwidth}{!}{%
\begin{tikzpicture}[
    >=Latex,
    font=\small,
    block/.style={
        draw,
        rounded corners=3pt,
        thick,
        align=center,
        inner sep=4pt,
        minimum height=0.5cm,
        minimum width=1.5cm
    },
    smallblock/.style={
        draw,
        rounded corners=3pt,
        thick,
        align=center,
        inner sep=4pt,
        minimum height=0.85cm,
        minimum width=2.25cm
    },
    decision/.style={
        draw,
        diamond,
        aspect=2.0,
        thick,
        align=center,
        inner sep=1pt,
        minimum width=2.3cm,
        minimum height=1.1cm
    },
    arrow/.style={->, thick},
    dasharrow/.style={->, thick, dashed}
]


\node[block, fill=gray!8] (home) at (-1.1,3.2)
{Camera node\\(RGB-D)};

\node[block, fill=cyan!8] (vision) at (0,1.3)
{Vision node \\(3D targets streaming, $\mathbf{p}_{cut}^{cam}$)};

\node[block, fill=green!8] (hrvstCntrol) at (4.3,1.3) 
{Harvest controller\\(task sequencing)};

\node[block, fill=green!8] (map) at (4.6,3.2)
{Robot-frame\\target mapping\\
$\mathbf{p}^{T} = \mathcal{T}_{cr}(\mathbf{p}_{cut}^{cam}) + \Delta\mathbf{p}_{d}$};

\node[block, fill=purple!8] (preposition) at (8.5,3.2)
{Target-conditioned\\pre-positioning\\
$\mathbf{q}^{ss}=f_{ss}(\mathbf{p}^{T})$};

\node[smallblock, dashed, fill=white] (safety) at (13.7,3.4)
{Hardware safety wrapper\\
(time shaping, accel. checks, HOLD/HALT)};

\node[block, fill=blue!6] (PPO) at (8.55,1.3)
{PSA-PPO local reaching\\
$\bar{\mathbf{u}}_t = \mu_{\theta}(\boldsymbol{\xi}_t)$};

\node[block, fill=blue!6] (streamer) at (12.5,1.3)
{Policy interface\\
(action processing \\+ joint mapping)};

\node[block, rotate=90,  fill=teal!8] (robot) at (16.1,.3)
{UR10e control interface};

\node[block, fill=orange!10] (grip) at (4.3,-.4)
{Gripper controller};


\draw[arrow] (home.south) -- ($(vision.north)-(11mm, 0)$);
\draw[arrow] (vision) -- (hrvstCntrol);
\draw[arrow] (hrvstCntrol) -- ($(map.south)-(3mm, 0)$);

\draw[arrow] (map) -- (preposition);

\draw[arrow] (preposition) -- (PPO);

\draw[arrow] (PPO) -- (streamer);

\draw[arrow] (streamer) -- ($(robot.north)+(0, 10mm)$);

\draw[arrow] (hrvstCntrol) --  (grip);
\draw[arrow] (grip.east) -- ($(robot.north)-(0, 7mm)$);;

\draw[dasharrow] ($(safety.south)+(21mm, 0)$) to[bend right=15] ($(robot.east)-(2mm, 0)$);
\draw[dasharrow] ($(robot.east)+(2mm, 0)$) to[bend right=15] ($(safety.south)+(25mm, 0)$);


\begin{pgfonlayer}{background}
\node[
    draw,
    dashed,
    thick,
    rounded corners=5pt,
    fill=gray!5,
    inner xsep=0.3cm,
    inner ysep=0.3cm,
    fit=(home)(vision)(hrvstCntrol)(map)(preposition)(PPO)(streamer)(robot)(grip)(safety)
] (rosbox) {};
\end{pgfonlayer}

\node[
    anchor=south west,
    font=\footnotesize\bfseries,
    inner sep=2pt
] at ([xshift=0.15cm,yshift=0.15cm]rosbox.south west)
{ROS-based harvesting control};

\end{tikzpicture}%
}
\caption{ROS-based harvesting-control architecture. Smoothed RGB-D cutting points are transformed to the robot frame and corrected for end-effector geometry. The harvest controller performs target-conditioned pre-positioning before activating the PSA-PPO policy for the learned local side approach. Policy outputs are mapped to UR10e joint commands through the ROS trajectory interface, with independent hardware-side timing and acceleration safeguards. A separate gripper-control branch actuates the 2FG7-driven cut-and-hold mechanism and releases harvested lettuce at the collection basket.}
\label{fig:ros_harvesting_control}
\end{figure*}
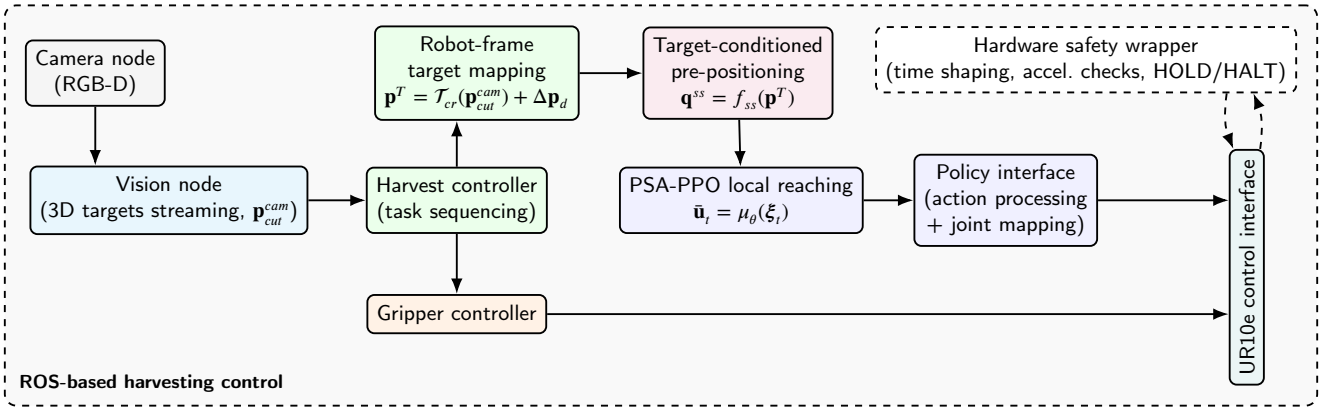

The selected PSA-PPO actor was integrated with the ROS harvesting architecture shown in Figure~\ref{fig:ros_harvesting_control}. Simulation validation used randomly generated workspace targets, whereas physical harvesting received smoothed cutting points from the RGB-D perception module (Sec. \ref{subsec:vision_module}); the target-conditioned geometry and learned local-reaching formulation were otherwise retained during deployment.

At the beginning of a harvesting sequence, the UR10e remained at its home configuration while available lettuce targets were localized. Each camera-frame target $\mathbf{p}_{cut}^{cam}$ was transformed to the robot frame and corrected for the cut-and-hold tool geometry:
\begin{equation}
\mathbf{p}^{T}
=
\mathcal{T}_{cr}
\left(
\mathbf{p}_{cut}^{cam}
\right)
+
\Delta\mathbf{p}_{d}
\label{eq:deployment_target_mapping}
\end{equation}
\noindent where $\mathcal{T}_{cr}(\cdot)$ denotes the calibrated rigid camera-to-robot transformation and $\Delta\mathbf{p}_{d}=[\Delta x_d,\Delta y_d,\Delta z_d]^\top$ denotes the fixed deployment correction accounting for residual perception, calibration, and end-effector geometry. The correction was determined before the reported evaluation and was subsequently held fixed.

No external sequence of intermediate side-approach waypoints was generated during deployment. Instead, $\mathbf{p}^{T}$ defined the same target-conditioned approach geometry used during training. The arm first moved to the corresponding deterministic safe-start configuration, after which PSA-PPO controlled the final local side approach; $\mathbf{p}^{S}$ therefore remained an internal geometric reference rather than a separately commanded physical waypoint.

During inference, the actor was evaluated deterministically:
\begin{equation}
\bar{\mathbf{u}}_t
=
\mu_{\theta}
\left(
\boldsymbol{\xi}_t
\right),
\qquad
\mathbf{q}^{ROS}_{cmd,t}
=
f_{map}
\left(
\mathbf{q}^{cmd}_t
\right)
\label{eq:deployment_policy_mapping}
\end{equation}
\noindent where $\mu_{\theta}(\cdot)$ is the deterministic actor output, $\mathbf{q}^{cmd}_t$ follows Eq.~\eqref{eq:lettuce_action_processing}, and $f_{map}(\cdot)$ maps the simulation joint ordering to the physical UR10e controller interface.

The ROS interface updated the policy state from UR10e joint feedback and published single-point \texttt{JointTrajectory} commands to the scaled joint-trajectory controller. For each command, feasible per-joint durations were computed from the current state, commanded displacement, and prescribed velocity and acceleration limits, and their maximum was used as the synchronized trajectory duration. Acceleration was checked before command publication and from measured joint-velocity feedback; exceeding the corresponding safety threshold replaced the pending motion with a hold command and entered a halted state. These execution safeguards operated outside the learned task objective.

After the target-reaching condition was satisfied, the harvest controller closed the OnRobot 2FG7 gripper to actuate the cut-and-hold mechanism, transferred the harvested lettuce to the collection basket, and reopened the gripper for release. When multiple targets were available, the remaining target with the smallest Euclidean distance from the current TCP position was selected; otherwise, the robot returned home.

\section{Experiment}
\label{experiment}

\subsection{Experimental Setup}
\label{subsec:experimental_platform}

PSA-PPO training and simulation evaluation were conducted using NVIDIA Isaac Lab in Isaac Sim 5.1.0 with the RSL-RL backend. Training used GPU-parallel simulation on an Ubuntu 22.04 LTS workstation equipped with dual Intel Xeon Gold processors, 256~GB RAM, and an NVIDIA RTX A6000 GPU with 48~GB VRAM. Candidate policies were trained without physical robot interaction, and the selected actor was evaluated deterministically on a fixed set of randomly generated targets spanning the prescribed front workspace. The component-removal ablations were evaluated on the same targets, enabling paired target-wise comparison with the complete PSA-PPO formulation.

Physical experiments used an RB-VOGUI mobile platform (Robotnik Automation, Paterna, Spain) equipped with a UR10e manipulator (Universal Robots, Odense, Denmark), an OnRobot 2FG7 gripper (OnRobot, Odense, Denmark), and an Intel\textsuperscript{\textregistered} RealSense\textsuperscript{\texttrademark}
D435i RGB-D camera mounted near the tool center point. The adapted end effector was mounted directly to the 2FG7, while an onboard Intel i7-based computer executed perception, target transformation, harvesting sequencing, deterministic PSA-PPO inference, ROS~1 communication, and gripper control. The simulation-trained actor was transferred to this platform without any fine-tuning.

Two complementary physical experiments were conducted. First, a controlled motion benchmark compared PSA-PPO with Cartesian+TOTG, RRTConnect, and RRT* over the same scheduled target locations, isolating controller behavior from perception and crop-interaction effects. Second, the complete PSA-PPO system was evaluated on greenhouse lettuce harvesting trials at the Texas A\&M AgriLife Research Center in Dallas, TX, USA, where cutting-point targets were generated online by the RGB-D perception module. However, calibration was performed to map the targets between the camera and robot frames. Thus, the greenhouse experiment evaluated the complete vision-to-harvest pipeline. Planner configurations and benchmark implementation details are reported in Appendix~\ref{app:C}.

\subsection{Evaluation Protocol}
\label{subsec:evaluation_metrics}

\begin{table*}
\centering
\small
\caption{Evaluation protocol used for simulation, controlled physical
benchmarking, and greenhouse harvesting.}
\label{tab:lettuce_eval_protocol}
\begin{tabular}{@{}p{3.0cm}p{3.3cm}p{9.0cm}@{}}
\toprule
\textbf{Evaluation} &
\textbf{Population} &
\textbf{Primary outcomes} \\
\midrule

Simulation validation &
200 workspace targets &
Reaching the target w/o any safety measures, path-safe reaching, strict success, final TCP error, and
maximum below-plane violation \\

Formulation ablation &
Same 200 targets per formulation &
Reach without safety measures, path-safe, and strict success; target-frame behavior, maximum
below-plane violation, trajectory length, and episode duration \\

Physical benchmark &
150 scheduled targets per controller; $N=97$ common-success targets for
continuous comparisons &
Target-reaching reliability, endpoint error, local TCP acceleration and jerk,
joint-position tracking error, normalized target convergence, and trajectory
geometry \\

Greenhouse harvesting &
144 harvesting attempts &
Target reaching, cutting, retention, release, and complete harvesting success \\

\bottomrule
\end{tabular}
\end{table*}

Evaluation was organized at three levels: simulation, controlled physical
motion, and greenhouse harvesting, with the formulation ablation embedded
within the simulation evaluation. Table~\ref{tab:lettuce_eval_protocol}
summarizes the corresponding populations and primary outcomes.

For simulation target $k$, the terminal TCP error was:
\begin{equation}
d_k^{T}
=
\left\|
\mathbf{p}^{T}_k
-
\mathbf{p}^{tcp}_{k,f}
\right\|_2
\label{eq:eval_final_target_error}
\end{equation}
Let $I_{k}^{\mathrm{final}}$ indicate satisfaction of the final capture
criterion over the prescribed validation hold interval, and let
$I_{k}^{\mathrm{safe}}$ indicate satisfaction of both the safe-start and
learned-path tolerances,
\begin{equation}
I_{k}^{\mathrm{safe}}
=
\mathbb{I}
\left[
b_{0,k}\leq\varepsilon_{\mathrm{safe}}
\;\land\;
b_{k}^{\max}\leq\varepsilon_{\mathrm{safe}}
\right]
\label{eq:eval_path_safe_indicator}
\end{equation}
Strict success required both conditions, such that
$I_{k}^{\mathrm{strict}}
=
I_{k}^{\mathrm{final}}I_{k}^{\mathrm{safe}}$.
The corresponding simulation rates were:
\begin{equation}
R_{r}
=
\frac{100}{N_v}
\sum_{k=1}^{N_v}
I_{k}^{r},
\quad
r\in
\left\{
\mathrm{final},
\mathrm{safe},
\mathrm{strict}
\right\}
\label{eq:eval_sim_success}
\end{equation}
For each component-removal formulation $a$, the paired strict-success loss relative to the complete PSA-PPO formulation was defined over the identical target set as:
\begin{equation}
\Delta R_{\mathrm{strict}}^{(a)}
=
\frac{100}{N_v}
\sum_{k=1}^{N_v}
\left(
I_{k,\mathrm{full}}^{\mathrm{strict}}
-
I_{k,a}^{\mathrm{strict}}
\right)
\label{eq:eval_ablation_delta}
\end{equation}
Target-frame errors, tool orientation, maximum below-plane violation,
trajectory length, episode duration, and motion quantities were retained for
the behavioral analyses reported with the ablation results.

Let $\mathcal{M} = \{\mathrm{PSA\mbox{-}PPO}, \mathrm{Cartesian+TOTG}, \mathrm{RRTConnect}, \linebreak \mathrm{RRT^*}\} $
denote the set of controllers for the controlled physical benchmark. Physical reaching for target $k$ and controller $c$ was defined by-
\begin{equation}
I_{k,c}^{\mathrm{reach}}
=
\mathbb{I}
\left[
d_{k,c}^{T}\leq\varepsilon_r
\right],
\quad
R_{\mathrm{reach}}^{(c)}
=
\frac{100}{N_b}
\sum_{k=1}^{N_b}
I_{k,c}^{\mathrm{reach}}
\label{eq:eval_physical_reach}
\end{equation}
where $\varepsilon_r=0.020$~m was the common physical evaluation tolerance.
Continuous controller comparisons were restricted to the mutually successful
target set:
\begin{equation}
\mathcal{K}_{\cap}
=
\bigcap_{c\in\mathcal{M}}
\left\{
k:
I_{k,c}^{\mathrm{reach}}=1
\right\},
\quad
N_{\cap}
=
\left|
\mathcal{K}_{\cap}
\right|
\label{eq:eval_common_success}
\end{equation}
TCP transforms and joint states were recorded at 100~Hz, uniformly resampled,
and smoothed using a 0.15-s moving-average window before derivative-based
analysis. With $a_i=\|\mathbf{a}_i\|_2$ and
$j_i=\|\mathbf{j}_i\|_2$, motion intensity was quantified by-
\begin{equation}
A_{\mathrm{rms}}
=
\sqrt{
\frac{1}{N_a}
\sum_{i=1}^{N_a}a_i^2
},
\quad
J_{\mathrm{rms}}
=
\sqrt{
\frac{1}{N_j}
\sum_{i=1}^{N_j}j_i^2
}
\label{eq:eval_rms_motion}
\end{equation}
together with the 99.5th-percentile jerk $J_{99.5}$. The primary local-motion
comparison used the final 0.22~m of target-relative approach. Executed
trajectory length and geometric detour were defined as:
\begin{equation}
L_{\mathrm{path}}
=
\sum_{m=1}^{M-1}
\left\|
\mathbf{s}_{m+1}-\mathbf{s}_{m}
\right\|_2,
\quad
D_{\mathrm{ratio}}
=
\frac{
L_{\mathrm{path}}
}{
\left\|
\mathbf{s}_{M}-\mathbf{s}_{1}
\right\|_2
}
\label{eq:eval_path_metrics}
\end{equation}
where $D_{\mathrm{ratio}}\geq1$, with values closer to unity indicating more
direct motion. Endpoint error, joint-position tracking RMSE, normalized
target convergence, and additional trajectory-geometry measures were
evaluated from the same paired trajectories. Detailed trajectory processing,
planner settings, and benchmark-integrity procedures are provided in
Appendix~\ref{app:C}.

For greenhouse attempt $k$, harvesting progression was evaluated sequentially
through reaching, cutting, retention, and release. If $N_m$ denotes the number
of attempts completing stage $m$, the cumulative and conditional stage rates
were:
\begin{equation}
\begin{split}
& R_m
=
\frac{N_m}{N_h}\times100\%,
\quad
R_{m\mid m-1}
=
\frac{N_m}{N_{m-1}}\times100\%\\
& m\in
\left\{
\mathrm{reach},
\mathrm{cut},
\mathrm{hold},
\mathrm{release}
\right\}
\end{split}
\label{eq:eval_stage_success}
\end{equation}
where the conditional rate was applied from the cutting stage onward. Because
successful release required completion of all preceding stages, overall
harvesting success was:
\begin{equation}
H_{\mathrm{success}}
=
R_{\mathrm{release}}
=
\frac{N_{\mathrm{complete}}}{N_h}\times100\%
\label{eq:eval_harvest_success}
\end{equation}
Failures were assigned to the first unsuccessful harvesting stage. Paired
statistical tests, bootstrap confidence intervals, multiplicity correction,
and sensitivity analyses used for the simulation ablation and physical
benchmark are detailed in Appendices~\ref{app:B} and~\ref{app:C}.

\subsection{Results}
\label{sub:results}

\subsubsection{Simulation Validation and Formulation Ablation}
\label{subsubsec:simulation_results}

\begin{table*}
\centering
\small
\caption{Component-removal ablation of PSA-PPO over the same 200 simulation
targets. Continuous quantities are reported as median [Q1,Q3].}
\label{tab:psa_ppo_ablation_main}
\begin{tabular}{@{}lcccc@{}}
\toprule
\textbf{Metric} &
\textbf{PSA-PPO} &
\textbf{w/o path safety} &
\textbf{w/o side corridor} &
\textbf{w/o motion reg.} \\
\midrule

Final reach (\%) &
99.0 & 89.0 & \textbf{100.0} & 98.0 \\

Path-safe (\%) &
\textbf{92.5} & 69.5 & 73.0 & 88.5 \\

Strict success (\%) &
\textbf{92.0} & 63.5 & 73.0 & 87.0 \\

Final TCP error (mm) &
32.3 [29.2, 35.1] &
32.1 [28.3, 38.3] &
\textbf{25.3 [21.4, 28.6]} &
35.0 [30.5, 39.2] \\

P95 maximum below-plane violation (mm) &
\textbf{2.11} & 26.8 & 14.3 & 4.52 \\

Trajectory length (m) &
\textbf{0.406 [0.264, 0.547]} &
0.443 [0.294, 0.567] &
0.461 [0.294, 0.592] &
0.464 [0.309, 0.592] \\

Duration (s) &
1.47 [0.933, 2.24] &
\textbf{1.13 [0.867, 1.60]} &
1.40 [0.967, 2.04] &
1.52 [1.13, 3.12] \\

\bottomrule
\end{tabular}

\vspace{2pt}
\begin{minipage}{0.98\textwidth}
\footnotesize
One independently trained checkpoint was evaluated for each formulation.
Boldface identifies the numerically favorable value for each row and does not
imply overall controller superiority. Path-safe and strict criteria were
evaluated identically for all formulations.
\end{minipage}

\end{table*}

\begin{figure*}
\centering
\subfloat[]{
    \includegraphics[width=0.47\textwidth]
    {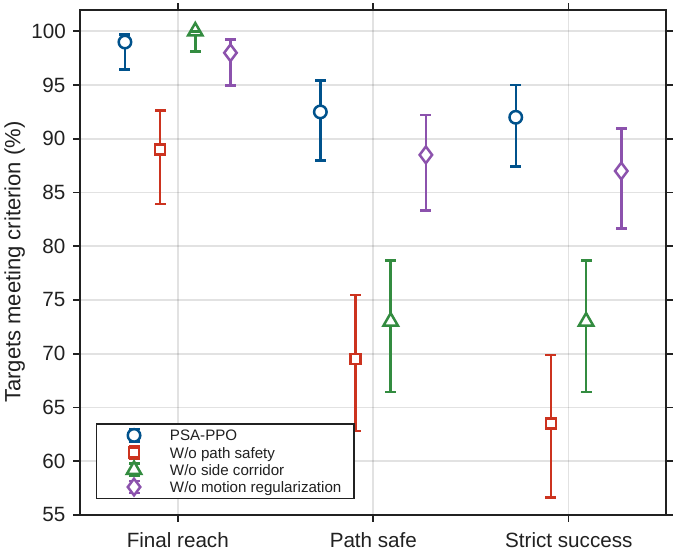}
    \label{fig:sim_ablation_success}}
\hfill
\subfloat[]{
    \includegraphics[width=0.47\textwidth]
    {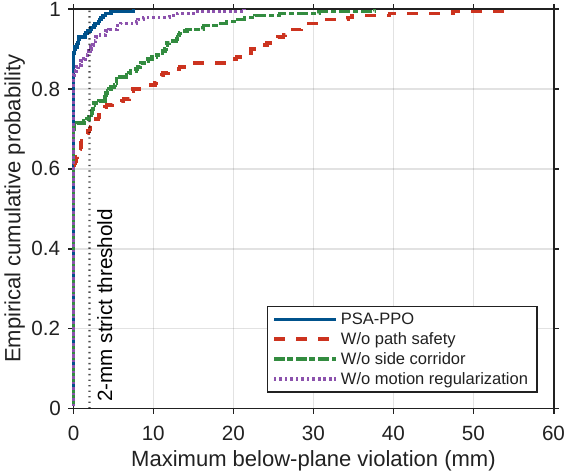}
    \label{fig:sim_ablation_safety_ecdf}}
\caption{Effect of PSA-PPO component removal on harvesting-compatible
simulation performance over the same 200 targets.
(a) Final reaching, path-safe reaching, and strict success with Wilson
95\% confidence intervals.
(b) Empirical distribution of the maximum wrist--TCP below-plane
violation; the dashed vertical line denotes the 2-mm strict path-safety
threshold.}
\label{fig:sim_ablation_outcomes}
\end{figure*}

\noindent\textbf{Simulation validation and component effects.}
The selected PSA-PPO policy satisfied the final capture criterion for
198 of 200 targets (99.0\%), the path-safety criterion for 185 trajectories
(92.5\%), and both conditions for 184 targets, yielding 92.0\% strict success.
The 7-percentage-point difference between final reaching and strict success
was dominated by trajectory-level clearance: among the 16 strict failures,
14 were safety-only cases, one was a capture-only case, and one failed both
criteria. Thus, terminal target acquisition alone did not fully characterize
the harvesting-compatible behavior captured by the strict criterion.

The component-removal results in Table~\ref{tab:psa_ppo_ablation_main}
show that path-safety shaping had the largest association with strict
performance. Removing these terms reduced final reaching from 99.0\% to
89.0\%, path-safe success from 92.5\% to 69.5\%, and strict success from
92.0\% to 63.5\%. On the paired target set, PSA-PPO remained strictly
successful for 61 targets on which the no-path-safety formulation failed,
whereas the reverse occurred for only four targets. The resulting
strict-success difference was 28.5 percentage points
(95\% CI: 21.5--35.5 percentage points;
Holm-adjusted $p=1.18\times10^{-13}$). The P95 maximum below-plane
violation increased from 2.11 to 26.8~mm, and the largest observed violation
increased from 9.68 to 54.3~mm.

Removing the side-corridor terms produced a different trade-off. This
formulation reached all 200 targets and achieved a smaller median final TCP
error than PSA-PPO (25.3 versus 32.3~mm), yet path-safe and strict success
both decreased to 73.0\%. The paired strict-success loss was
19.0 percentage points
(95\% CI: 13.5--25.0 percentage points;
Holm-adjusted $p=8.22\times10^{-10}$). The result therefore separates
terminal precision from the compound harvesting criterion: improved endpoint
accuracy did not imply improved path-safe reaching.

Removal of motion regularization produced the smallest task-level effect.
Final reaching remained 98.0\%, while path-safe and strict success decreased
to 88.5\% and 87.0\%, respectively. The paired strict-success loss was
5.0 percentage points
(95\% CI: 1.5--8.5 percentage points;
Holm-adjusted $p=0.0129$). However, the corresponding raw-action-rate,
action-acceleration, and TCP-jerk diagnostics did not increase monotonically
when regularization was removed. The evaluated checkpoint therefore does not
support attributing lower jerk directly to the regularization terms; instead,
their measurable task-level association was a modest increase in strict
success together with a different trajectory-duration trade-off.

Figure~\ref{fig:sim_ablation_outcomes} provides the corresponding population-level view. In addition to the strict-success degradation, the below plane empirical cumulative distribution function (ECDF) shows that removal of the path-safety formulation substantially increased the upper tail of trajectory-level violation.

\begin{figure*}[!t]
\centering
\includegraphics[width=0.95\textwidth]
{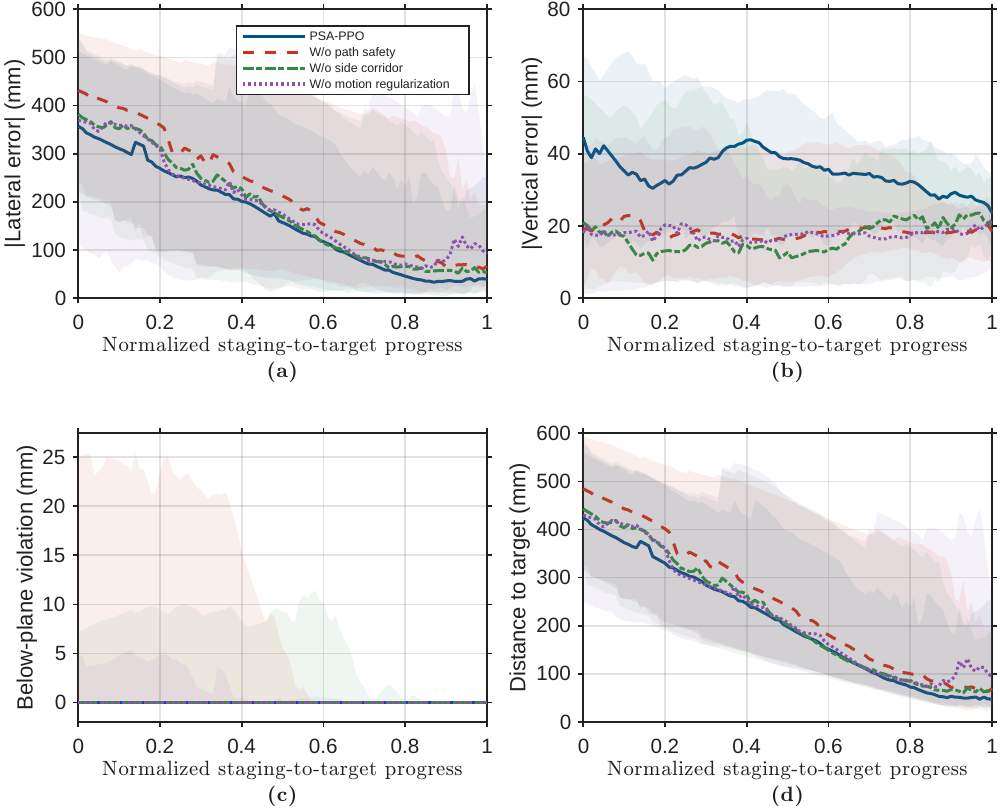}
\caption{Target-frame trajectory profiles for PSA-PPO and the three
component-removal formulations over the common 200-target evaluation set:
(a) absolute lateral error, (b) absolute vertical error,
(c) wrist--TCP below-plane violation, and (d) TCP distance to the target
as functions of normalized staging-to-target progress. Curves denote median
behavior and shaded regions the corresponding trajectory dispersion.}
\label{fig:sim_ablation_profiles}
\end{figure*}

\noindent\textbf{Behavioral and workspace robustness.}
The target-frame profiles in Figure~\ref{fig:sim_ablation_profiles} show
that the component removals altered different aspects of the local approach
rather than uniformly scaling trajectory quality. The clearest separation
occurred in the path-safety channel, where the no-path-safety formulation
produced substantially larger below-plane excursions. In contrast, the
lateral and vertical profiles did not exhibit a consistent ordering across
all formulations. This behavior is consistent with the side-corridor terms
acting within a coupled target-closure and constraint-shaping objective,
rather than as an independent fixed-band controller. All formulations reduced
TCP-to-target distance toward the end of the approach, but differed in whether
that convergence was achieved while satisfying the complete strict criterion.

\begin{figure*}
\centering
\includegraphics[width=0.95\textwidth]
{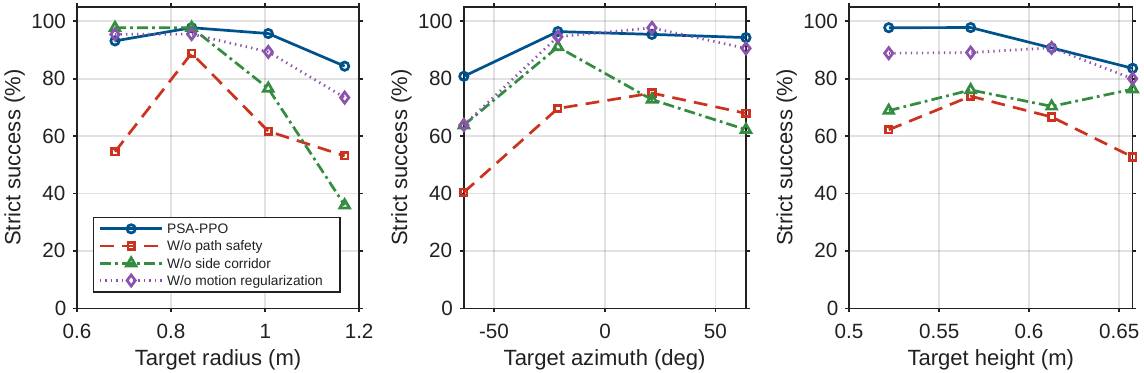}
\caption{Strict-success performance of PSA-PPO and the three
component-removal formulations across common bins of
target radius, target azimuth, and target height (left to right).
The bin-wise results characterize variation over the evaluated target set
for the selected checkpoints and not variability across independent policy
retraining.}
\label{fig:sim_ablation_workspace_bin}
\end{figure*}

The workspace analysis in Figure~\ref{fig:sim_ablation_workspace_bin}
shows that PSA-PPO retained comparatively high strict success across the
evaluated radius, azimuth, and height bins, whereas removal of the path-safety
or side-corridor terms produced larger degradation in several regions of the
workspace. The motion-regularization formulation remained closer to PSA-PPO
throughout most bins, consistent with its smaller aggregate strict-success
difference.

Taken together, the simulation results identify path-safety shaping as the
largest task-level contributor among the tested removals and show that
side-corridor shaping can improve strict harvesting-compatible performance
even when terminal accuracy alone improves after its removal. Detailed
paired outcomes, fixed-band corridor diagnostics, orientation and
motion-regularization analyses, three-dimensional workspace maps, and
representative trajectories are provided in Appendix~\ref{app:B}. Because
one independently trained checkpoint was evaluated per formulation, these
comparisons characterize the evaluated policies over the common target set
and do not quantify retraining variability.

\subsubsection{PSA-PPO vs. Classical Motion Planners on the Physical Robot}
\label{subsubsec:inhouse_controller_comparison}

\noindent\textbf{Task-level reliability, terminal precision, and failure behavior.}
PSA-PPO achieved the highest observed target-reaching rate, successfully
reaching 142 of 150 targets (94.7\%), compared with 132/150 (88.0\%) for
RRTConnect, 129/150 (86.0\%) for Cartesian+TOTG, and 120/150 (80.0\%) for
RRT* (Table~\ref{tab:real_reliability}). Thus, the simulation-trained policy
retained high task-level reliability after transfer to the physical UR10e.

\begin{table}
\centering
\caption{Physical target-reaching reliability of PSA-PPO and the classical
planning baselines under the common 20-mm reaching criterion. Here, CI denotes the confidence interval.}
\label{tab:real_reliability}
\small
\setlength{\tabcolsep}{4.0pt}
\renewcommand{\arraystretch}{1.12}
\begin{tabular}{@{}lccc@{}}
\toprule
\textbf{Controller} &
\textbf{Reached} &
\textbf{Failed} &
\textbf{Success (\%) [95\% CI]} \\
\midrule
PSA-PPO            & \textbf{142/150} & \textbf{8}  & \textbf{94.7 [89.8, 97.3]} \\
Cartesian+TOTG     & 129/150          & 21          & 86.0 [79.5, 90.7] \\
RRTConnect         & 132/150          & 18          & 88.0 [81.8, 92.3] \\
RRT*               & 120/150          & 30          & 80.0 [72.9, 85.6] \\
\bottomrule
\end{tabular}
\end{table}

\begin{figure}
\centering
\includegraphics[width=0.95\linewidth]
{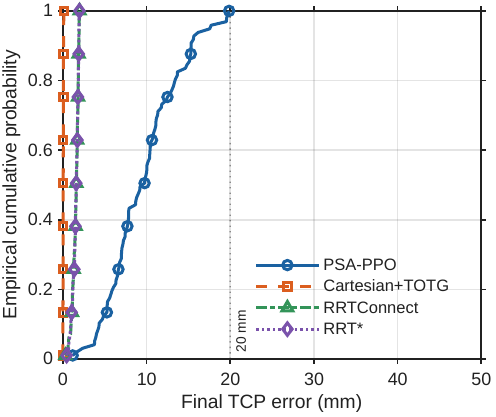}
\caption{Empirical cumulative distribution of physical TCP
endpoint error. The vertical dashed line denotes the 20-mm
reaching criterion. Planner-based control produced tighter
endpoint distributions, whereas PSA-PPO exhibited a broader
but predominantly tolerance-compliant distribution.}
\label{fig:real_endpoint_ecdf}
\end{figure}

On the common-success targets, the median final TCP error was
9.78~mm for PSA-PPO, compared with 0.06~mm for Cartesian+TOTG,
1.62~mm for RRTConnect, and 1.61~mm for RRT*. The endpoint ECDF in
Figure~\ref{fig:real_endpoint_ecdf} shows this separation across the
distribution. The classical planners therefore retained a substantial
advantage in terminal positioning precision, while the PSA-PPO median
remained below one-half of the 20-mm task-level tolerance. However, while the planner-based approach struggled to reach all targets, PSA-PPO outperformed it by 94.7\%, achieving success while maintaining all safety criteria.  

\begin{figure}
\centering
\includegraphics[width=0.95\linewidth]
{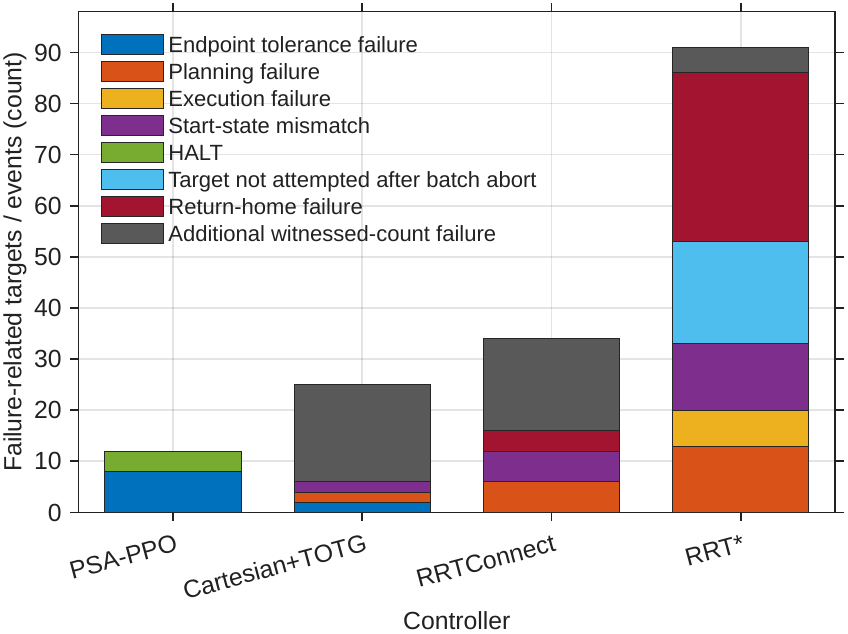}
\caption{Recorded failure-related events during the physical
controller benchmark. Cell counts denote diagnostic events
and are not necessarily mutually exclusive; the corresponding
numbers of unique unsuccessful targets were 8 for PSA-PPO,
21 for Cartesian+TOTG, 18 for RRTConnect, and 30 for RRT*.}
\label{fig:real_failure_modes}
\end{figure}

Failure diagnostics further distinguish the controllers beyond their binary
success rates. PSA-PPO unsuccessful runs were concentrated more strongly in
target-reaching events, whereas the planner executions exhibited a broader
mixture of planning-, execution-, return-, and controller-related diagnostic
events (Figure~\ref{fig:real_failure_modes}). Because these event categories
were not mutually exclusive, the figure is interpreted independently of the
unique unsuccessful-target counts in Table~\ref{tab:real_reliability}.

\noindent\textbf{Local motion dynamics and target convergence.}
Within the final 0.22-m target-relative approach, PSA-PPO produced median
RMS acceleration and jerk of 0.127~m/s$^2$ and 11.38~m/s$^3$,
respectively, compared with 0.528 and 42.17 for Cartesian+TOTG,
0.555 and 40.74 for RRTConnect, and 0.267 and 18.91 for RRT*
(Table~\ref{tab:real_local_motion}). Relative to the planner medians,
the PSA-PPO reductions in RMS jerk were 73.0\%, 72.1\%, and 39.8\%.
The corresponding reductions in RMS acceleration were 75.9\%, 77.1\%,
and 52.4\%. High-end transients followed the same ordering:
the 99.5th-percentile jerk was 77.72~m/s$^3$ for PSA-PPO versus
131.88, 164.77, and 127.84~m/s$^3$, respectively. Joint-position tracking
RMSE was also lowest for PSA-PPO. All PSA-PPO--planner paired contrasts for
the four local-motion metrics remained significant after Holm correction
(adjusted $p=4.02\times10^{-16}$).

\begin{table*}
\centering
\caption{Physical motion quality within the final 0.22-m target-relative
approach. Values are medians over the common-success target set ($N=97$).
Lower values indicate lower dynamic excitation or tracking error.}
\label{tab:real_local_motion}
\small
\setlength{\tabcolsep}{8pt}
\renewcommand{\arraystretch}{1.15}
\begin{tabular}{@{}lcccc@{}}
\toprule
\textbf{Metric} &
\textbf{PSA-PPO} &
\textbf{Cartesian+TOTG} &
\textbf{RRTConnect} &
\textbf{RRT*} \\
\midrule

RMS acceleration (m/s$^{2}$)
& \textbf{0.1270} & 0.5277 & 0.5549 & 0.2668 \\

RMS jerk (m/s$^{3}$)
& \textbf{11.38} & 42.17 & 40.74 & 18.91 \\

99.5th-percentile jerk (m/s$^{3}$)
& \textbf{77.72} & 131.88 & 164.77 & 127.84 \\

Joint-position tracking RMSE (rad)
& $\boldsymbol{6.96\times 10^{-4}}$
& $2.56\times10^{-3}$
& $2.47\times10^{-3}$
& $1.15\times10^{-3}$ \\

\bottomrule
\end{tabular}
\end{table*}

\begin{figure*}
\centering
\subfloat[]{
\includegraphics[width=0.475\textwidth]
{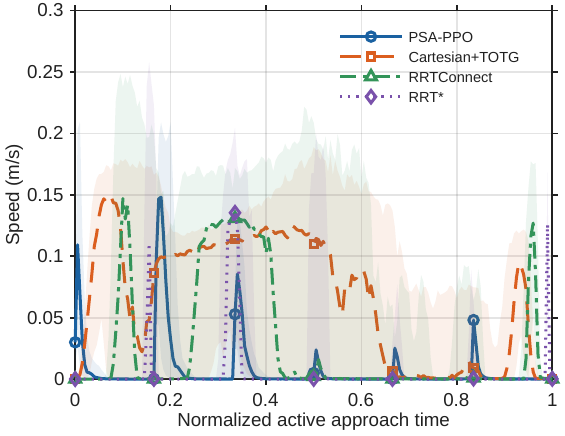}
\label{fig:real_profile_speed}}
\hfill
\subfloat[]{
\includegraphics[width=0.475\textwidth]
{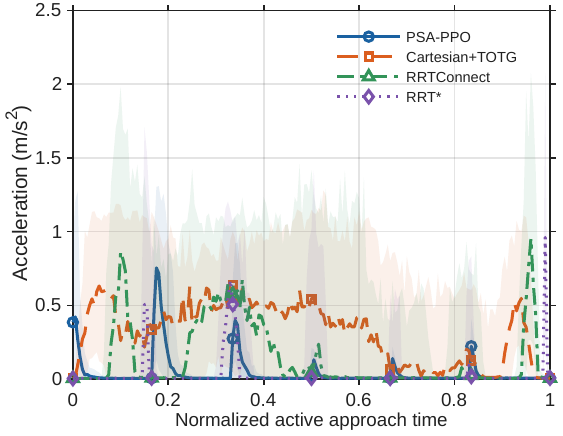}
\label{fig:real_profile_acc}}
\\[2mm]
\subfloat[]{
\includegraphics[width=0.475\textwidth]
{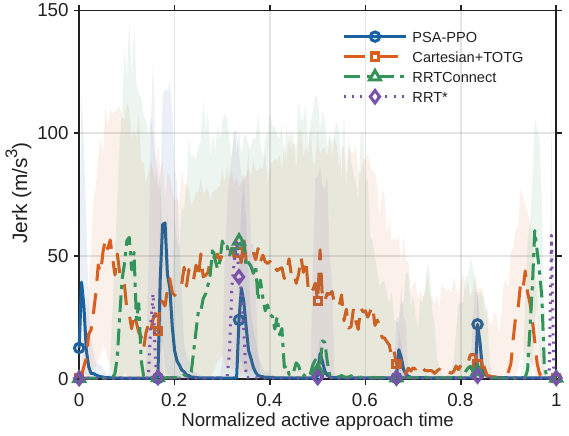}
\label{fig:real_profile_jerk}}
\hfill
\subfloat[]{
\includegraphics[width=0.475\textwidth]
{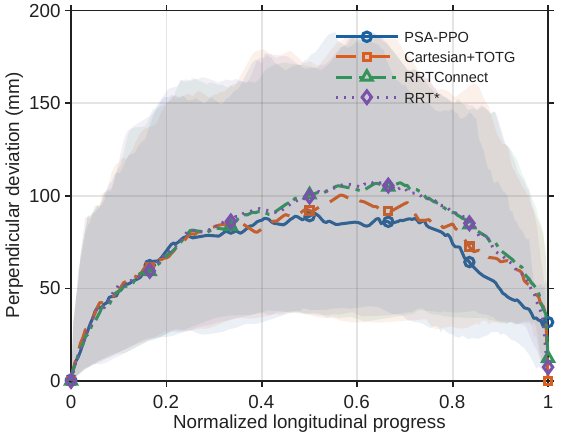}
\label{fig:real_profile_dev}}

\caption{Measured trajectory profiles for the four physical controllers over
the common-success target population: (a) TCP speed, (b) acceleration,
(c) jerk over normalized active approach progression, and (d) perpendicular
deviation from the start--target reference line over normalized longitudinal
progress. The normalized axes describe trajectory progression rather than
absolute wall-clock execution time.}
\label{fig:real_motion_profiles}
\end{figure*}

The normalized motion profiles in Figure~\ref{fig:real_motion_profiles}
show that the lower aggregate PSA-PPO acceleration and jerk were sustained
over substantial portions of the local approach rather than arising solely
from isolated endpoints. The perpendicular-deviation profiles, however,
overlap more strongly across controllers. Lower dynamic excitation should
therefore not be interpreted as evidence that PSA-PPO followed the most
geometrically direct Cartesian path.

\begin{figure}
\centering
\includegraphics[width=0.95\linewidth]
{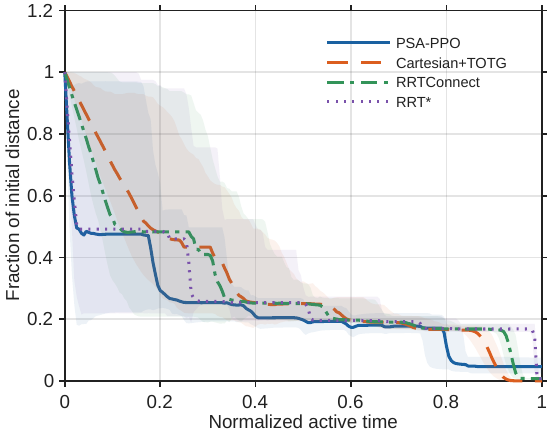}
\caption{Normalized target-distance contraction during the final 0.22-m
approach over the common-success target set. Remaining TCP-to-target distance
is normalized by its value at local-window entry, while the horizontal axis
represents each trajectory's normalized active progression. PSA-PPO reached
50\% remaining distance earlier in normalized progression than the three
planner-based controllers.}
\label{fig:real_target_convergence}
\end{figure}

Target-distance contraction also differed across controllers. PSA-PPO reduced
the remaining target distance to 50\% of its local-window entry value after
a median normalized progression of 0.195, compared with 0.405 for
Cartesian+TOTG, 0.370 for RRTConnect, and 0.345 for RRT*
(Figure~\ref{fig:real_target_convergence}). PSA-PPO therefore performed a
larger fraction of its target closure earlier in its own normalized local
approach. This quantity characterizes convergence shape and does not represent
wall-clock execution speed.

\noindent\textbf{Trajectory geometry and controller trade-off.}
Lower PSA-PPO dynamic excitation did not correspond to shorter or more direct
paths. Over the complete reconstructed approach, the median detour ratio was
1.81 for PSA-PPO, compared with 1.67 for Cartesian+TOTG, 1.68 for
RRTConnect, and 1.70 for RRT*. PSA-PPO also accumulated greater total joint
travel than the planner-based methods. The population-level geometry therefore
favored the classical planners in path directness, while the physical
motion-quality results favored PSA-PPO in acceleration, jerk, and tracking
error. Detailed trajectory-geometry statistics, reliability geometry
trade-offs, and matched examples of atypically expansive planner trajectories
are provided in Appendix~\ref{app:C}.

Overall, the physical benchmark reveals a controller trade-off rather than
uniform dominance: the classical planners achieved substantially greater
terminal precision and more direct average trajectories, whereas PSA-PPO
achieved the highest observed reaching rate together with markedly lower
local dynamic excitation and joint-tracking error.

\subsubsection{Target-Reaching Performance in Harvesting}
\label{subsubsec:greenhouse_harvesting}
The end-effector of the integrated system successfully reached 136 out of 144 vision-derived lettuce cutting points, achieving a target success rate of 94.4\% in the greenhouse (Table~\ref{tab:greenhouse_harvest_performance}). Cutting was completed successfully for 124 plants, which accounts for 86.1\% of all attempts after reaching the target. Additionally, we noted that 13 plants were not properly secured after cutting, and 3 plants fell during transfer to the basket.


\begin{table}
\centering
\caption{Stage-wise performance of the integrated robotic lettuce harvesting
system under greenhouse conditions.}
\label{tab:greenhouse_harvest_performance}
\small
\setlength{\tabcolsep}{4.5pt}
\renewcommand{\arraystretch}{1.12}
\begin{tabular}{@{}lccc@{}}
\toprule
\textbf{Harvesting stage} &
\textbf{Successful} &
\textbf{Overall (\%)} &
\textbf{Conditional (\%)} \\
\midrule
Target reached     & 136/144 & 94.4 & 94.4 \\
Lettuce cut        & 124/144 & 86.1 & 91.2 \\
\bottomrule
\end{tabular}

\vspace{2pt}
\parbox{0.97\columnwidth}{\footnotesize
Overall percentages use all 144 attempts as the denominator. Conditional
percentages are evaluated relative to successful completion of the immediately
preceding stage.}
\end{table}

However, the conditional rates reveal that the most harvesting failures occur after target acquisition. Cutting succeeded for 124 of the 136 reached plants, resulting in a success rate of 91.2\%. Retention succeeded for 111 of 124 cut plants (89.5\%), and basket release
succeeded for 108 of 111 retained plants, achieving a success rate of 97.3\%. Therefore, the principal loss in complete-system performance occurred during physical interaction between the crop and the end-effector, rather than motion policy control.

\begin{figure}
\centering
\includegraphics[width=0.97\linewidth]{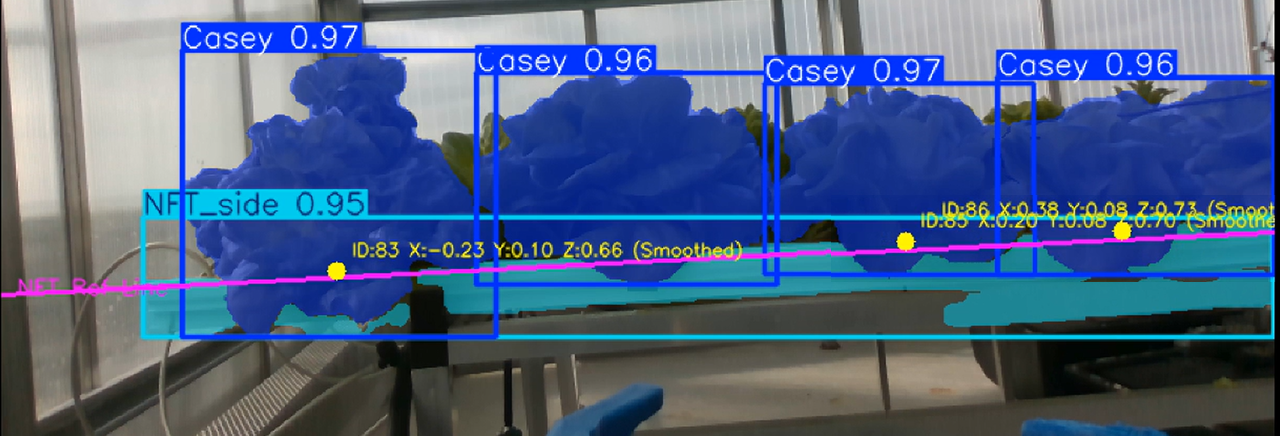}
\caption{Representative online target stream during greenhouse harvesting. The image was acquired from the robot-mounted RGB-D camera while the UR10e remained at the home configuration. The vision node identified the available lettuce targets and published their corresponding 3D cutting-point coordinates for subsequent robot-frame transformation and target-conditioned reaching.}
\label{fig:greenhouse_target_stream}
\end{figure}

\begin{figure*}
\centering
\makebox[\textwidth][c]{%
\begin{minipage}[t]{0.46\textwidth}
\vspace{0pt}
\centering
\begin{subfigure}[t]{\linewidth}
    \centering
    \includegraphics[width=\linewidth,keepaspectratio]
    {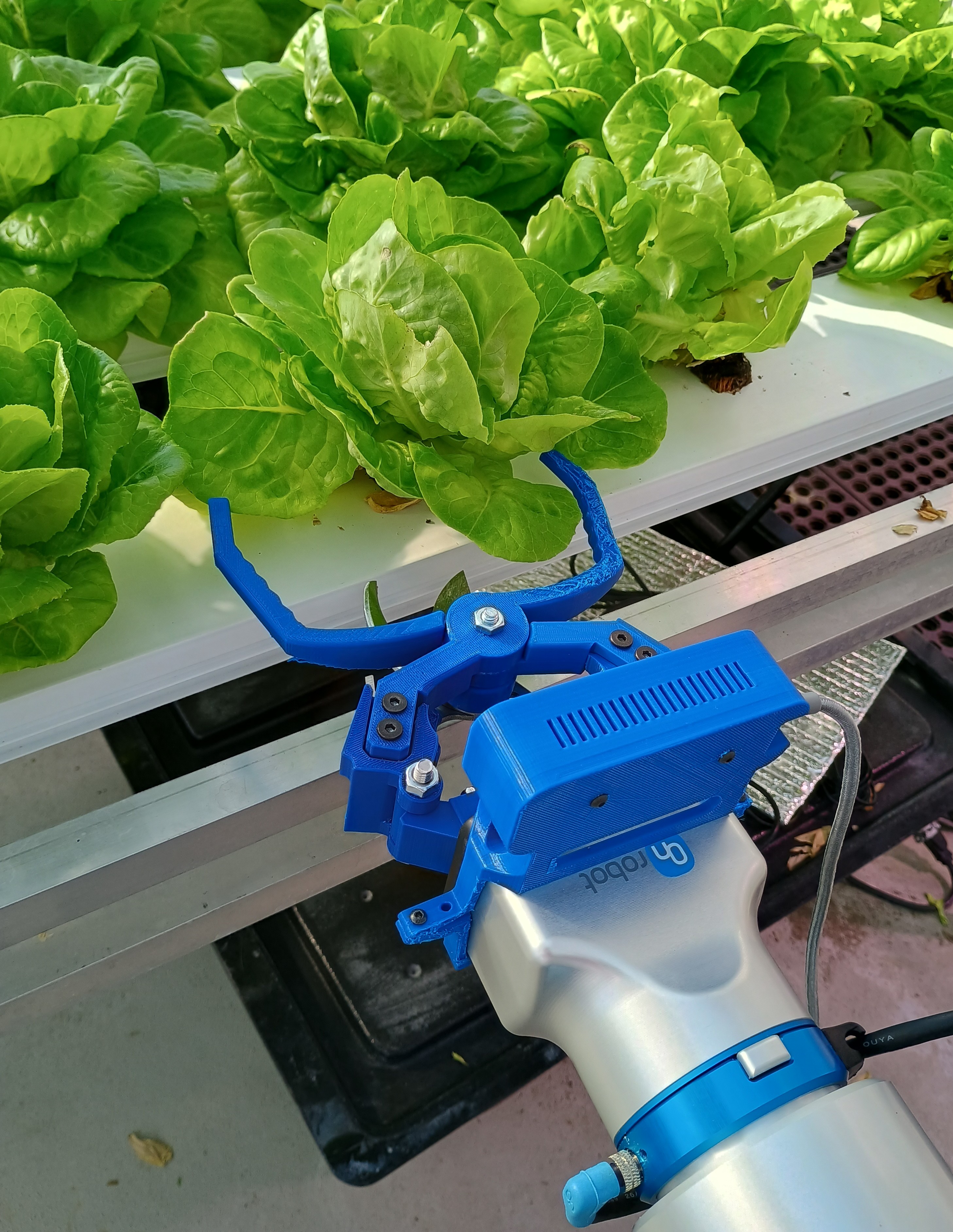}
    \caption{}
    \label{fig:greenhouse_approach}
\end{subfigure}
\end{minipage}%
\hspace{0.008\textwidth}%
\begin{minipage}[t]{0.39\textwidth}
\vspace{0pt}
\centering
\begin{subfigure}[t]{\linewidth}
    \centering
    \includegraphics[width=\linewidth,keepaspectratio]
    {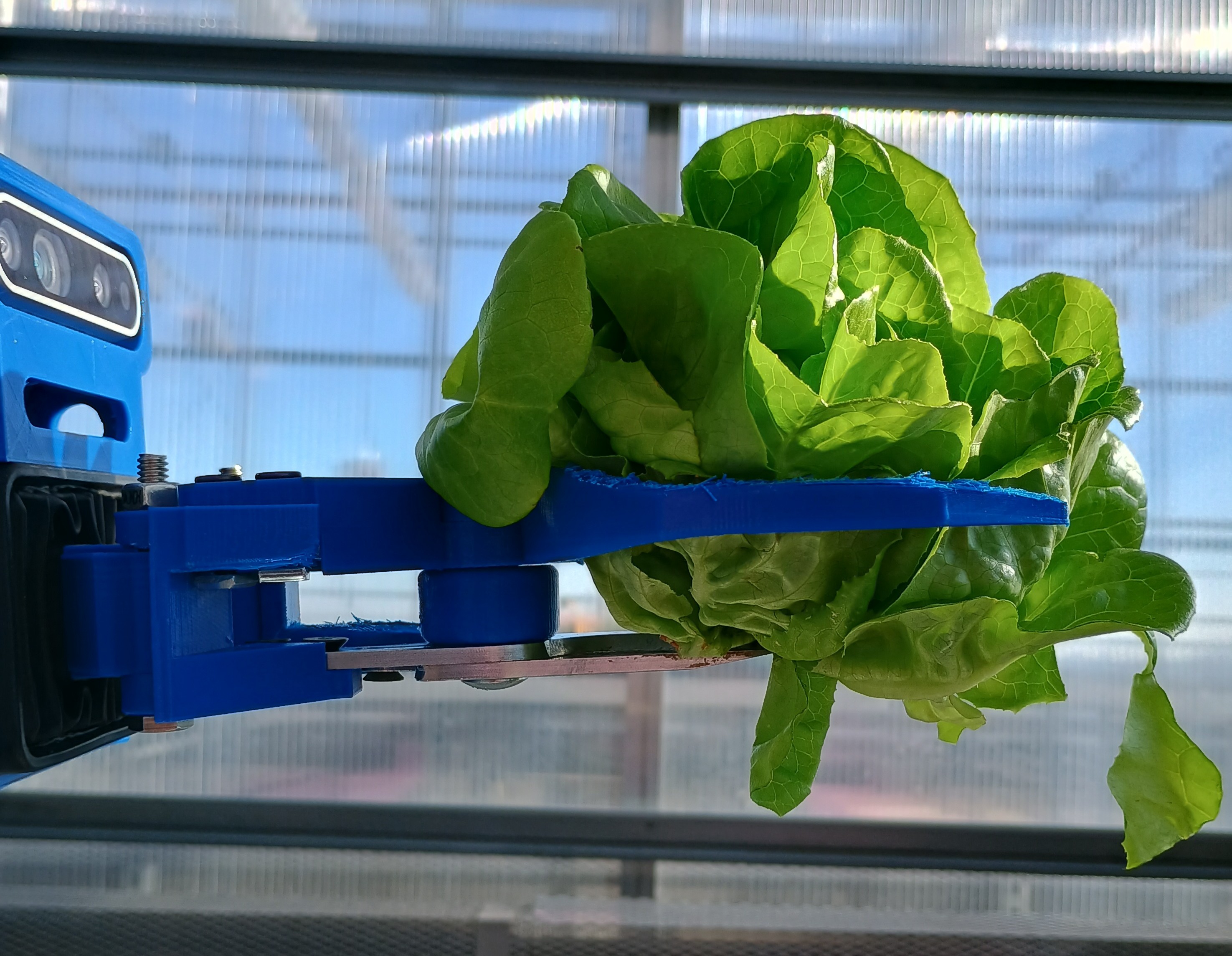}
    \caption{}
    \label{fig:greenhouse_cut_hold}
\end{subfigure}

\vspace{0.008\textwidth}

\begin{subfigure}[t]{\linewidth}
    \centering
    \includegraphics[width=\linewidth,keepaspectratio]
    {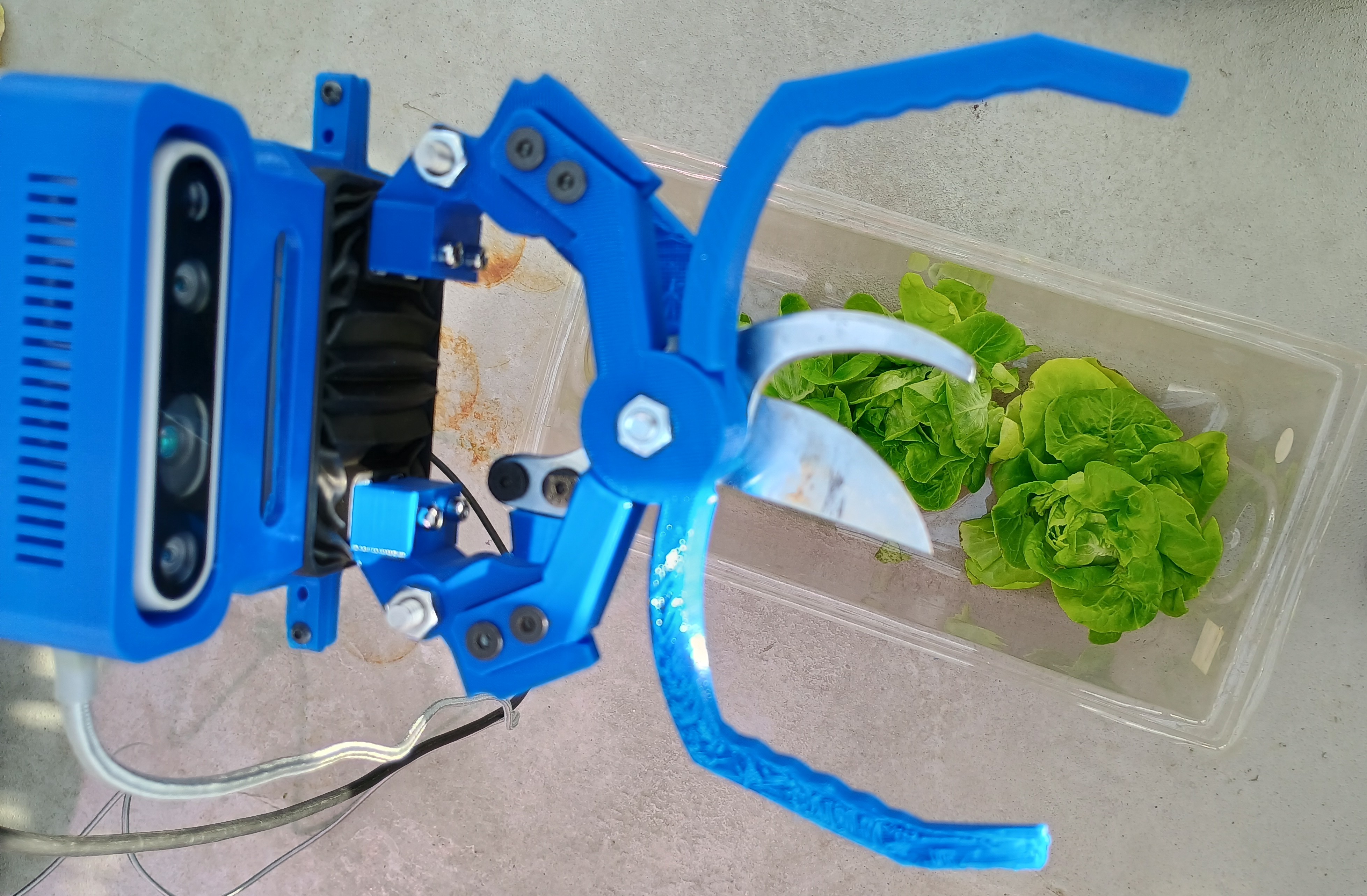}
    \caption{}
    \label{fig:greenhouse_release}
\end{subfigure}
\end{minipage}%
}

\caption{Representative greenhouse execution of the integrated harvesting
pipeline: (a) PSA-PPO-controlled side approach toward the vision-derived
cutting location; (b) simultaneous cutting and retention using the
2FG7-driven cut-and-hold end effector; and (c) release of the harvested
lettuce into the collection basket.}
\label{fig:greenhouse_harvest_sequence}
\end{figure*}

Figure~\ref{fig:greenhouse_target_stream} shows the online target stream generated by the vision module, while Figure~\ref{fig:greenhouse_harvest_sequence} depicts a typical execution of the resulting physical sequence: PSA-PPO performed the
vision-conditioned side approach, closure of the 2FG7-actuated cutting and retention, and the retained lettuce was subsequently transferred and released into the collection basket.

The greenhouse evaluation therefore shows that the simulation-trained
PSA-PPO controller retained a high reaching rate when driven by online RGB-D
targets within the complete harvesting system, while the principal remaining
performance losses occurred downstream during cutting and retention.

\subsection{Discussion}
\label{subsec:discussion}

The simulation results indicate that the main value of the PSA-PPO
formulation lies in how harvesting-compatible success is structured rather
than in terminal reaching alone. Removing the path-safety terms produced the
largest degradation in strict success and substantially increased the tail of
below-plane violations, confirming that transient trajectory behavior can
remain unacceptable even when the final target is reached. The side-corridor
ablation provides an equally important counter-example: it improved terminal
accuracy and reached every target, yet reduced strict success considerably.
Thus, optimization toward the endpoint alone can produce a numerically better
reaching result while weakening the combined harvesting constraint. The
motion-regularization ablation was less conclusive; although its removal
reduced strict success, several action- and TCP-derivative measures did not
worsen. The present evidence therefore supports path-safety and corridor
shaping as important elements of the coupled task formulation, but does not
justify attributing lower jerk directly to the regularization term. This
task-structured use of simulation-trained control is related to recent
plant-manipulation work by Subedi et al.~\cite{Subedi2026FindFruit}, who
separated structured kinematic motion from a policy learned entirely in
simulation; however, their objective was occlusion removal rather than
harvesting-compatible crop-proximal reaching. Moreover, because one
independently trained checkpoint was evaluated per formulation, the ablation
establishes differences over the common target set rather than retraining
robustness.

\begin{figure*}
\centering
\subfloat[]{
    \includegraphics[width=0.47\textwidth]
    {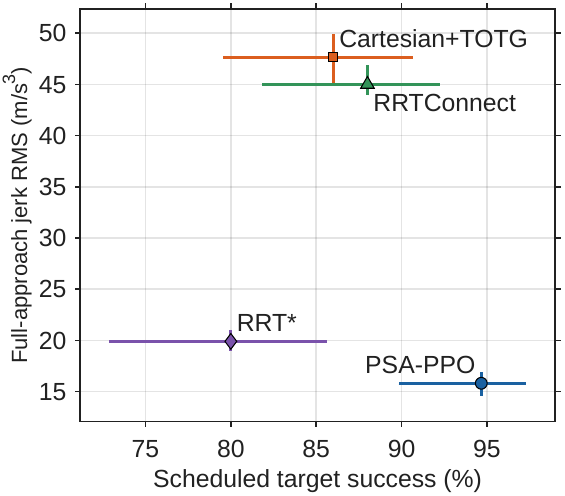}
    \label{fig:discussion_reliability_jerk}}
\hfill
\subfloat[]{
    \includegraphics[width=0.47\textwidth]
    {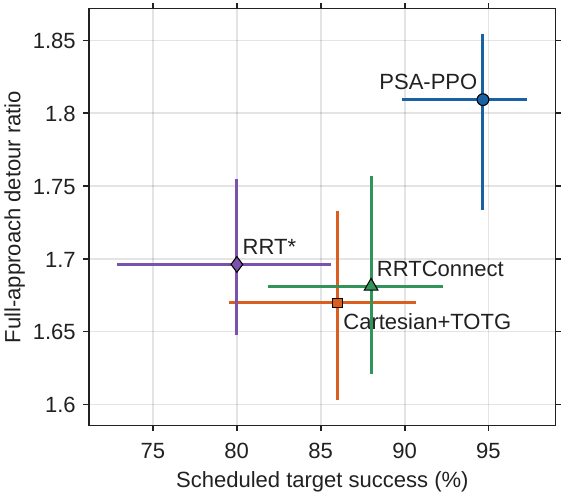}
    \label{fig:discussion_reliability_detour}}
\caption{Physical controller trade-offs between target-reaching reliability
and (a) median RMS jerk within the final 0.22-m approach and
(b) median trajectory detour ratio. PSA-PPO favors the
reliability--dynamic-excitation operating regime, whereas the planner-based
controllers retain greater geometric directness.}
\label{fig:discussion_controller_tradeoff}
\end{figure*}

The physical benchmark provides stronger evidence for transfer, while also
revealing a clear learned-versus-planned trade-off. PSA-PPO achieved the highest reaching reliability and
substantially lower local acceleration, jerk, and joint-tracking error, but
the classical planners retained markedly better endpoint precision and more
direct trajectories. These results are therefore complementary rather than
hierarchical: PSA-PPO occupied a favorable reliability--dynamic-excitation
regime, whereas the planners remained stronger in geometric precision and
path efficiency. Figure~\ref{fig:discussion_controller_tradeoff} visualizes
this trade-off directly. Recent agricultural planning work illustrates the value of
that geometric objective; for example, Kang et al.~\cite{Kang2025BroccoliPlanning}
explicitly optimized planning time, path length, and trajectory smoothness
for selective broccoli harvesting. The present results therefore do not
suggest replacing geometric planning where path optimality is primary, but
rather identify a different operating regime for learned crop-proximal
motion. This distinction is especially relevant near crops and rigid cultivation infrastructure, where smooth execution is desirable, but lower jerk is not itself evidence of collision safety. Similarly, occasional expansive planner trajectories should not override the population-level finding that their paths were more direct on average. The measured physical dynamics must also be interpreted as properties of the deployed controller stack, because PSA-PPO inference was combined with action filtering and an independent hardware safety wrapper rather than executed as an isolated learned policy.

The jerk reduction achieved by PSA-PPO relative to the classical planners has two separable contributors that should be distinguished. First, the deployment stack applies a temporal action filter ($\alpha = 0.18$) and an independent hardware acceleration safeguard (0.0045 rad/$s^2$) that attenuate high-frequency motion regardless of policy behavior; these are stack-level properties rather than learned ones, and their individual contribution to the observed jerk advantage cannot be isolated without a dedicated ablation. Second, Figure \ref{fig:real_motion_profiles} shows that lower acceleration and jerk are sustained over the full local approach rather than at isolated endpoints, consistent with the policy having learned intrinsically smoother convergence as an emergent property of the coupled task formulation. The motion-regularization ablation does not support attributing this smoothness directly to the regularization reward terms, since their removal did not produce a monotonic increase in raw-action diagnostics (Figure \ref{fig:sim_ablation_safety_ecdf}). Fully disaggregating these two contributors is left for future work.

The greenhouse experiment shifts the interpretation from controller behavior to system bottlenecks. Reaching remained high when predefined Cartesian targets were replaced by online RGB-D cutting points, suggesting that the sim-to-real policy did not undergo a large degradation after integration with perception, coordinate transformation, and greenhouse variability. The controlled and greenhouse populations are different; however, their similar reaching rates should not be treated as evidence of statistical equivalence. More importantly, incomplete harvests occurred after successful target reaching. The dominant residual losses therefore moved downstream to tool and crop interaction rather than target acquisition or basket release. This result both supports the usefulness of the learned motion controller and limits its system-level significance: improving reaching alone is unlikely to raise complete harvesting success substantially unless crop interaction is improved concurrently. Recent compliant-control work points to the same systems issue from a different direction; Wang et al.~\cite{WANG2025126} used adaptive variable impedance control and force sensing to regulate physical crop interaction during dragon-fruit picking. For the present system, this suggests that contact-aware control and retention mechanics may now offer greater system-level benefit than further reduction of terminal reaching error alone. Similar stage-dependent losses have also been observed in field-tested harvesting systems~\cite{Arad2020Sweeper}.

Several limitations consequently define the scope of the present findings. The term ``path-safe'' refers to the implementation with side-corridor entry, collision avoidance with wrist--TCP target-height clearance criteria, and does not provide formal uncertainty-awareness guarantees for the complete robot or greenhouse scene. The physical study used one manipulator, one greenhouse environment, a frozen policy deployment correction, and a single trained policy for each formulation. Consequently, the reported results characterize the evaluated policies but do not quantify variability across independent training runs. Broader claims therefore require repeated training seeds, additional robot and crop configurations, and evaluation under changing canopy and infrastructure geometry. In addition, the planner baselines were evaluated only in the controlled reaching benchmark and not as complete vision-to-harvest systems, while the perception and cutting modules were inherited from earlier work rather than independently re-evaluated here. Future research will investigate hybrid planning and control architectures that combine global scene-aware planning, learned crop-proximal control, and contact-aware crop interaction. Such a hybrid design
could preserve the planners' geometric strengths while exploiting the
reliability and lower local dynamic excitation observed for PSA-PPO.

\section{Conclusion}
\label{sec:conclusion}

This study formulated robotic lettuce harvesting as a target-conditioned,
path-safe, side-corridor approach problem and developed PSA-PPO for simulation training and physical deployment on a UR10e robot. PSA-PPO demonstrated that such a harvesting-specific policy can be optimized entirely in simulation and transferred to a physical robot without real-world policy fine-tuning. The formulation ablation further showed that terminal accuracy alone is an incomplete measure of harvesting-compatible motion: path-safety and side-approach constraints materially affected strict task performance even when endpoint reaching remained high. On the physical robot, PSA-PPO achieved the highest reaching reliability (94.7\%) and lower local dynamic excitation, while the classical planners retained advantages in terminal precision but remained unreliable in constrained conditions. The resulting contribution is therefore not universal controller superiority, but a distinct and practically useful operating regime for crop-proximal manipulation. Future work will integrate global geometric planning, learned crop-proximal control, explicit clearance modeling, and contact-aware feedback, while expanding validation across crop conditions and greenhouse environments.

\section*{Code Availability}
The source code and supporting materials will be made
available upon publication at
\url{https://github.com/bashir-tamu/PSA-PPO-Lettuce-Harvesting}.

\section*{Acknowledgments}
This research is partially supported by the United States Department of Agriculture (USDA)'s National Institute of Food and Agriculture (NIFA) Research Capacity Fund Hatch Program: TEX09954 (Accession No. 7002248) and Research Capacity Fund Multistate Hatch Program: TEX0-1-9916 (Accession No. 7008389). This publication is also supported by the Texas A\&M University Division of Research Targeted Proposal Teams (TPT) funding program and Texas A\&M AgriLife Research. Any opinions, findings, conclusions, or recommendations expressed in this publication are those of the authors and should not be construed to represent any official USDA or U.S. Government determination or policy. 

\printcredits

\bibliographystyle{cas-model2-names}

\bibliography{source}

\clearpage
\appendix
\counterwithin{table}{section}
\counterwithin{figure}{section}
\counterwithin{equation}{section}

\begin{center}
    {\Large\bfseries Appendix}
\end{center}
\setlength{\textfloatsep}{10pt plus 2pt minus 2pt}
\setlength{\floatsep}{10pt plus 2pt minus 2pt}
\setlength{\intextsep}{10pt plus 2pt minus 2pt}
\section{PSA-PPO Training and Task Configuration}
\label{app:A}

This appendix reports the numerical task and optimization settings used for
the final PSA-PPO policy. PSA-PPO used the standard clipped PPO update; the
name denotes the harvesting-specific initialization, observation, reward, and
strict-success formulation under which the policy was trained. The interaction
and success thresholds are listed in Table~\ref{tab:app_interaction}, the
training workspace and target-conditioned safe-start rule in
Table~\ref{tab:app_workspace}, the implemented reward terms in
Table~\ref{tab:app_reward}, and the PPO/network settings in
Table~\ref{tab:app_ppo}.

\begin{table}
\centering
\footnotesize
\caption{Policy interaction, action-processing, capture, and path-safety parameters.}
\label{tab:app_interaction}
\setlength{\tabcolsep}{2.5pt}
\renewcommand{\arraystretch}{1.06}
\begin{tabularx}{\columnwidth}{
    @{}
    >{\raggedright\arraybackslash}p{0.36\columnwidth}
    >{\centering\arraybackslash}p{0.19\columnwidth}
    >{\raggedright\arraybackslash}X
    @{}
}
\toprule
\textbf{Parameter} & \textbf{Value} & \textbf{Definition} \\
\midrule
Observation dimension & 42 & Policy input dimension \\
Action dimension & 6 & UR10e joint-space action dimension \\
Simulation step & $1/120$ s & Physics integration step \\
Control decimation & 4 & Physics steps per policy action \\
Policy frequency & 30 Hz & Effective policy rate \\
Episode limit & 9.0 s & Maximum training episode duration \\
TCP offset & 0.120 m & Tool offset along local $+Z$ \\
Action filter $\alpha$ & 0.18 & Temporal filter coefficient \\
Action scale $s_u$ & 0.065 & Joint-position increment scale \\
$\varepsilon_{ax}$ & 0.035 m & Axial capture tolerance \\
$\varepsilon_{lat}$ & 0.030 m & Lateral capture tolerance \\
$\varepsilon_z$ & 0.025 m & Vertical capture tolerance \\
$\varepsilon_d$ & 0.045 m & Euclidean target-distance tolerance \\
$\varepsilon_a$ & 0.95 & Minimum approach-axis alignment \\
$\varepsilon_v$ & 0.12 & Maximum vertical tool-axis component \\
Vertical soft threshold & 0.085 & Soft orientation-penalty threshold \\
Safety-plane margin & 0 m & Plane placed at target height \\
$\varepsilon_{safe}$ & 0.002 m & Maximum strict trajectory violation \\
$\varepsilon_{hard}$ & 0.050 m & Early-termination violation threshold \\
$N_s$ & 8 & Consecutive strict-capture steps \\
Final lateral band & 0.025 m & Lateral corridor tolerance \\
Final vertical band & 0.020 m & Vertical corridor tolerance \\
Final line band & 0.035 m & Perpendicular line tolerance \\
Staging radius & 0.090 m & Staging-neighborhood threshold \\
Corridor-state progress & $-0.05$ & Binary corridor-state threshold \\
\bottomrule
\end{tabularx}
\end{table}

The nominal target-conditioned safe-start posture was constructed from the
home configuration $\mathbf q^{home}$ and a geometry-dependent posture
$\mathbf q^{geo}$,
\begin{equation}
\begin{split}
&  \mathbf q^{ss}
=
(1-\beta)\mathbf q^{home}
+
\beta\mathbf q^{geo},\\
& \mathbf q^{home}
=
[-0.24,-1.57,1.57,0,1.57,0]^{\top}
\end{split}
\label{eq:app_safe_start_blend}
\end{equation}
where all joint angles are in radians and $\beta$ was selected from the target geometry. With target azimuth
$\phi=\operatorname{atan2}(p_y^T,p_x^T)$ in radians and
$z_n=\operatorname{clip}((p_z^T-0.59)/0.10,-1,1)$, the unblended posture was:
\begin{equation}
\mathbf q^{geo}
=
\begin{bmatrix}
\operatorname{clip}(-0.10+0.70\phi,-1.15,1.15)\\
-1.18-0.10z_n+\delta_s\\
1.24-0.04z_n+\delta_e\\
0.05\\1.57\\0
\end{bmatrix}
\label{eq:app_safe_start_geometry}
\end{equation}
During training, an independent uniform perturbation in
$[-0.004,0.004]$ rad was added to each joint after construction of the
nominal safe-start posture. Thus, the pre-positioning rule itself was
target-determined, while a small reset perturbation was retained during
policy training.

\begin{table*}
\centering
\footnotesize
\caption{Training-workspace sampling and target-conditioned safe-start configuration.}
\label{tab:app_workspace}
\setlength{\tabcolsep}{4.0pt}
\renewcommand{\arraystretch}{1.05}
\begin{tabularx}{\textwidth}{@{}p{3.3cm}p{4.1cm}X@{}}
\toprule
\textbf{Setting} & \textbf{Value} & \textbf{Implementation} \\
\midrule
\multicolumn{3}{@{}l}{\textit{Training-target distribution}} \\[1pt]
Target radius & $[0.60,1.25]$ m & Horizontal front-workspace radius \\
Target azimuth & $[-85^{\circ},85^{\circ}]$ & Front-workspace azimuth \\
Target height & $[0.50,0.68]$ m & Vertical target range \\
Approach distance $d_a$ & $[0.18,0.28]$ m & Same-height staging-to-target distance \\
Base radial sampling & Area-uniform & $\rho=\sqrt{\rho_{min}^2+u(\rho_{max}^2-\rho_{min}^2)}$ \\
Hard-region probability & 0.55 & A sampled target was reassigned to one of five hard-region modes with equal mode selection probability \\
Far-radius mode & $[1.08,1.25]$ m & Increased far-reach exposure \\
Negative-side mode & $\rho\in[1.08,1.25]$ m, \newline $\phi\in[-85^{\circ},-25^{\circ}]$ & Increased negative-lateral/corner exposure \\
Low-height mode & $\rho\in[0.90,1.25]$ m, \newline $z\in[0.50,0.585]$ m & Increased low/far exposure \\
High-height mode & $\rho\in[1.08,1.25]$ m, \newline $z\in[0.635,0.68]$ m & Increased high/far exposure \\
Side-boundary mode & $|\phi|\in[55^{\circ},85^{\circ}]$ & Increased left/right side-boundary exposure \\
\addlinespace
\multicolumn{3}{@{}l}{\textit{Geometry-aware safe-start selector}} \\[1pt]
Base blend $\beta$ & 0.82 & Default blend in Eq.~\eqref{eq:app_safe_start_blend} \\
Low-height rule & $z\leq0.58$ m $\Rightarrow\beta=0.76$ & Overrides stronger blending for low targets \\
Far-radius rule & $\rho\geq1.10$ m $\Rightarrow\beta\geq0.88$ & Applied only when the low-height rule is inactive \\
Very-far rule & $\rho\geq1.20$ m $\Rightarrow\beta\geq0.92$ & Applied only when the low-height rule is inactive \\
Negative/high rule & $p_y^T\leq-0.50$ m and $z\geq0.64$ m $\Rightarrow\beta\geq0.92$ & Additional negative-side/high-target blend \\
Far-negative/high rule & $p_y^T\leq-0.65$ m and $z\geq0.64$ m $\Rightarrow\beta\geq0.95$ & Strongest blend rule \\
Blend clamp & $[0.70,0.98]$ & Final admissible blend interval \\
Far-radius posture offset & $(\delta_s,\delta_e)=(+0.02,-0.03)$ rad & Shoulder/elbow offset for $\rho\geq1.10$ m \\
Far-negative/high offset & $(\delta_s,\delta_e)=(+0.04,-0.04)$ rad & Overrides the far-radius offset in the far-negative/high region \\
Training reset perturbation & $\pm0.004$ rad/joint & Uniform perturbation applied after nominal safe-start construction \\
\bottomrule
\end{tabularx}
\end{table*}

\begin{table*}
\centering
\scriptsize
\caption{Implementation-level reward terms used for the final PSA-PPO policy. The component symbols correspond to the compact decomposition in the main text.}
\label{tab:app_reward}
\setlength{\tabcolsep}{3.5pt}
\renewcommand{\arraystretch}{1.05}
\begin{tabularx}{\textwidth}{@{}p{2.0cm}p{5.1cm}p{3.3cm}X@{}}
\toprule
\textbf{Component} & \textbf{Implemented term} & \textbf{Weight} & \textbf{Definition / activation} \\
\midrule
$\mathcal R_t^{tar}$
& Staging-distance progress; staging distance
& $+25;\;-10$
& Multiplied by $1-\sigma(14(\ell_t-0.02))$ \\

$\mathcal R_t^{tar}$
& Target-distance progress; axial-error progress; line-progress increment; clipped line progress
& $+720;\;+610;\;+360;\;+185$
& Distance/axial increments clipped to $\pm0.04$; line increment to $\pm0.05$; direct $\ell_t$ term clipped to $[-0.3,1]$ \\

$\mathcal R_t^{tar}$
& $\exp(-7d_t^T)$; target shells at $d_t^T<0.15,0.10,0.07$ m
& $+145;\;+105;\;+165;\;+260$
& Near-target shaping \\

$\mathcal R_t^{tar}$
& Stop-short; staging loiter; target distance; $|e_t^{ax}|$; $|e_t^{lat}|$; $|e_t^z|$; axial overshoot
& $-950; \;-260; \;-190; \;-150;$\newline$-180; \;-260; \;-360$
& Staging loiter is active within 0.09 m of $\mathbf p^S$ when $d_t^T>0.10$ m; target-overshoot margin is 0.025 m \\

$\mathcal R_t^{cor}$
& Excessive TCP height; line-band violation; lateral-band violation; vertical-band violation; corridor axial overshoot
& $-260;\;-330;\;-310;\;-360;$\newline$-130$
& The first four terms use $\sigma(10(\ell_t+0.08))$ where applicable; excessive-height threshold is 0.030 m \\

$\mathcal R_t^{ori}$
& Tool-axis alignment reward; approach-axis deficit; vertical-orientation violation; soft vertical-orientation violation
& $+150;\;-1100;\;-780;\;-320$
& Alignment reward $\exp[-5(1-c_t^a)]$; hard/soft vertical thresholds are 0.12/0.085 \\

$\mathcal R_t^{safe}$
& Within-tolerance indicator; instantaneous below-plane violation; squared violation; accumulated path violation; initial violation
& $+80;\;-4200;\;-12000;\;-3600;$\newline$-2600$
& Uses instantaneous $b_t$, accumulated $\max_{\tau\le t}b_\tau$, and the initial safe-start violation \\

$\mathcal R_t^{reg}$
& Constant step cost; action magnitude; action-rate; action acceleration; joint velocity
& $-0.10;\;-0.0035;\;-0.015;$\newline$-0.026;\;-0.00050$
& Penalizes $\|\mathbf u_t\|_2^2$, first/second action differences, and $\|\dot{\mathbf q}_t\|_2^2$; no causal jerk interpretation is implied \\

$\mathcal B_t$
& Valid capture $\chi_t$; strict capture $\chi_t^{strict}$
& $+250;\;+2400$
& Strict capture additionally requires initial and accumulated path violation $\leq\varepsilon_{safe}$ \\
\bottomrule
\end{tabularx}
\end{table*}

The final-pose capture predicate required the axial, lateral, vertical, and
Euclidean position thresholds in Table~\ref{tab:app_interaction},
$c_t^a>\varepsilon_a$, $c_t^z<\varepsilon_v$, and positive instantaneous
wrist--TCP chain clearance. Strict capture additionally required both the
initial and accumulated below-plane violations to remain within
$\varepsilon_{safe}$. Successful termination required this strict predicate
for $N_s$ consecutive policy steps; an accumulated violation above
$\varepsilon_{hard}$ terminated the episode early.

\begin{table*}
\centering
\scriptsize
\caption{Exact training modifications used for the PSA-PPO component-removal ablations. Only the listed reward or termination logic was changed relative to the full formulation; observation dimensionality, action processing, target sampling, safe-start construction, capture thresholds, and the post-training strict evaluator were retained.}
\label{tab:app_ablation_definition}
\setlength{\tabcolsep}{4pt}
\renewcommand{\arraystretch}{1.08}
\begin{tabularx}{\textwidth}{@{}p{2.6cm}Xp{5cm}@{}}
\toprule
\textbf{Formulation} &
\textbf{Training modification relative to PSA-PPO} &
\textbf{Preserved elements} \\
\midrule

PSA-PPO &
No component removed. The $+2400$ strict-capture bonus required final capture, acceptable safe-start clearance, and acceptable accumulated path clearance. Successful termination used the same strict predicate, and accumulated below-plane violation above 0.050~m triggered hard-safety termination. &
Complete target, corridor, orientation, path-safety, and motion-regularization formulation. \\

w/o path safety (C2b) &
Removed the five path-safety reward terms:
$+80\,\mathbb{I}[b_t\leq\varepsilon_{\mathrm{safe}}]$,
$-4200\,b_t$,
$-12000\,b_t^2$,
$-3600\,\max(b_t^{\max}-\varepsilon_{\mathrm{safe}},0)$, and
$-2600\,b_0$.
The $+2400$ bonus remained, but its gate was changed from strict capture to final capture only.
Training success termination likewise used final capture only, and hard-safety termination was disabled. &
The 42-D observation, including wrist--TCP clearance channels, was unchanged.
Target/corridor/orientation rewards, motion regularization, safe-start logic, and path-safety logging were retained.
Strict path safety was restored only in the common post-training evaluator. \\

w/o side corridor (C2c) &
Removed staging/corridor shaping:
$+25$ staging-progress,
$-10$ staging-distance,
$+360$ line-progress increment,
$+185$ clipped line progress,
$-260$ staging loiter,
$-260$ excessive-TCP-height,
$-330$ line-band,
$-310$ lateral-band,
$-360$ vertical-band, and
$-130$ corridor axial-overshoot terms. &
Target-distance and target-frame error terms, orientation, path-safety rewards/history, strict-capture termination, hard-safety termination, and motion regularization were unchanged. \\

w/o motion regularization (C2d) &
Removed the four action/joint regularizers:
$-0.0035\|\mathbf{u}_t\|_2^2$,
$-0.015\|\Delta\mathbf{u}_t\|_2^2$,
$-0.026\|\Delta^2\mathbf{u}_t\|_2^2$, and
$-0.00050\|\dot{\mathbf{q}}_t\|_2^2$.
The constant per-step cost $-0.10$ was retained. &
Target, corridor, orientation, path-safety, strict-capture, and termination logic were unchanged. \\

\bottomrule
\end{tabularx}

\vspace{2pt}
\begin{minipage}{0.98\textwidth}
\footnotesize
All formulations were trained from scratch with seed 42 and the same actor--critic architecture and PPO optimizer settings. The full runner allowed 12,000 iterations, whereas the three ablation runners allowed 7,000; the reported random-200 evaluations used the model-6000 checkpoint for each formulation, providing the same evaluated optimization horizon. All four checkpoints were evaluated with the identical strict PSA-PPO validation criteria.
\end{minipage}
\end{table*}

\begin{table}
\centering
\footnotesize
\caption{PPO optimization and actor--critic network configuration.}
\label{tab:app_ppo}
\setlength{\tabcolsep}{2.5pt}
\renewcommand{\arraystretch}{1.07}
\begin{tabularx}{\columnwidth}{
    @{}
    >{\raggedright\arraybackslash}p{0.37\columnwidth}
    >{\centering\arraybackslash}p{0.22\columnwidth}
    >{\raggedright\arraybackslash}X
    @{}
}
\toprule
\textbf{Parameter} & \textbf{Value} & \textbf{Setting} \\
\midrule
Training seed & 42 & RSL-RL seed \\
Rollout length & 32 steps/env. & Samples per environment/update \\
Maximum iterations & 12,000 & PPO iterations \\
Actor hidden layers & 256--256--128 & Fully connected \\
Critic hidden layers & 256--256--128 & Fully connected \\
Activation & ELU & Hidden activation \\
Initial action-noise std. & 0.36 & Stochastic actor \\
Observation normalization & Disabled & Empirical normalization off \\
PPO clip $\epsilon$ & 0.20 & Ratio clipping \\
Clipped value loss & Enabled & Value-function clipping \\
Learning rate & $8\times10^{-5}$ & Initial rate \\
Learning-rate schedule & Adaptive & KL-based adaptation \\
Discount $\gamma$ & 0.99 & Return discount \\
GAE $\lambda$ & 0.95 & Advantage estimator \\
Entropy coefficient & 0.0015 & Exploration regularization \\
Value-loss coefficient & 1.0 & Critic weighting \\
Learning epochs & 5 & Epochs/update \\
Mini-batches & 4 & Mini-batches/update \\
Desired KL & 0.010 & Adaptive-schedule target \\
Maximum gradient norm & 1.0 & Gradient clipping \\
\bottomrule
\end{tabularx}
\end{table}

\FloatBarrier

\section{Simulation Validation and Formulation-Ablation Details}
\label{app:B}

This appendix details the fixed simulation-evaluation setup and diagnostics supporting Section~\ref{subsubsec:simulation_results}. Identical target coordinates and strict evaluation criteria were used for PSA-PPO and all component-removal baselines (Table~\ref{tab:app_sim_validation}).

\begin{table}
\centering
\footnotesize
\caption{Fixed random-workspace simulation-evaluation contract.}
\label{tab:app_sim_validation}
\setlength{\tabcolsep}{2.5pt}
\renewcommand{\arraystretch}{1.07}
\begin{tabularx}{\columnwidth}{
@{}
>{\raggedright\arraybackslash}p{0.39\columnwidth}
>{\centering\arraybackslash}p{0.24\columnwidth}
>{\raggedright\arraybackslash}X
@{}}
\toprule
\textbf{Parameter} & \textbf{Value} & \textbf{Setting} \\
\midrule
Targets per formulation & 200 & Identical target list \\
Target radius & $[0.60,1.25]$ m & Front workspace \\
Target azimuth & $[-85^\circ,85^\circ]$ & Front workspace \\
Target height & $[0.50,0.68]$ m & Vertical range \\
Radius sampling & Area-uniform & Horizontal annulus \\
Target-generation seed & 20260724 & Fixed evaluation set \\
Approach distance & 0.22 m & Same-height staging distance \\
Strict path tolerance & 0.002 m & Below-plane criterion \\
Safety-plane margin & 0 m & Target-height plane \\
Evaluation action multiplier & 0.75 & Deterministic evaluator \\
Near-target action multiplier & 0.40 & Applied within 0.10 m \\
Maximum policy steps & 280 & Per target \\
Additional geometric rules & Disabled & Evaluator-side rules off \\
\bottomrule
\end{tabularx}
\end{table}

Binary outcomes were paired by target. Risk differences were computed as
PSA-PPO minus the corresponding component-removal rate, with 95\% confidence
intervals from 10,000 paired bootstrap resamples. Exact McNemar tests were
used for paired binary outcomes and Holm correction was applied across the
three formulation contrasts within each outcome family. The target, rather
than trajectory samples, was the inferential unit. Because one independently
trained checkpoint was evaluated per formulation, these intervals and tests
characterize variation over the evaluation targets and not retraining
variability.

\begin{table*}
\centering
\footnotesize
\caption{Paired binary outcome differences between PSA-PPO and the three component-removal formulations over the common 200-target set. Positive $\Delta$ favors PSA-PPO.}
\label{tab:app_ablation_paired_binary}
\setlength{\tabcolsep}{4.5pt}
\renewcommand{\arraystretch}{1.07}
\begin{tabular}{@{}llcccc@{}}
\toprule
\textbf{Outcome} &
\textbf{Contrast} &
\textbf{PSA-PPO (\%)} &
\textbf{Variant (\%)} &
\textbf{$\Delta$ (pp) [95\% CI]} &
\textbf{Holm $p$} \\
\midrule
Final reach &
w/o path safety &
99.0 & 89.0 & 10.0 [6.0, 14.5] & $3.29\times10^{-5}$ \\
&
w/o side corridor &
99.0 & 100.0 & $-1.0$ [$-2.5$, 0.0] & 1.000 \\
&
w/o motion regularization &
99.0 & 98.0 & 1.0 [$-1.5$, 3.5] & 1.000 \\
\addlinespace

Path-safe &
w/o path safety &
92.5 & 69.5 & 23.0 [16.5, 29.5] & $1.14\times10^{-10}$ \\
&
w/o side corridor &
92.5 & 73.0 & 19.5 [14.0, 25.5] & $1.14\times10^{-10}$ \\
&
w/o motion regularization &
92.5 & 88.5 & 4.0 [1.0, 7.5] & 0.0386 \\
\addlinespace

Strict &
w/o path safety &
92.0 & 63.5 & 28.5 [21.5, 35.5] & $1.18\times10^{-13}$ \\
&
w/o side corridor &
92.0 & 73.0 & 19.0 [13.5, 25.0] & $8.22\times10^{-10}$ \\
&
w/o motion regularization &
92.0 & 87.0 & 5.0 [1.5, 8.5] & 0.0129 \\
\bottomrule
\end{tabular}
\end{table*}

\begin{figure}
\centering
\includegraphics[width=0.98\linewidth]{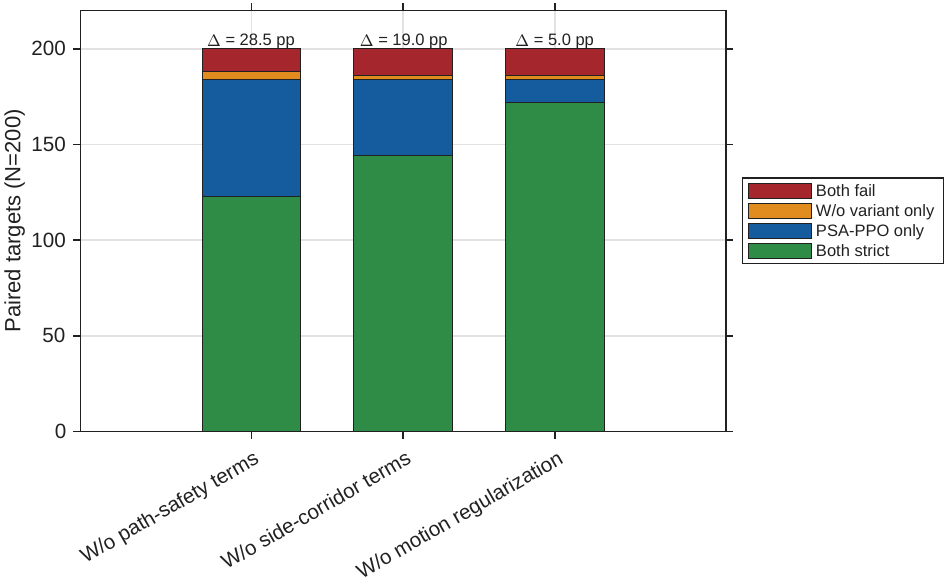}
\caption{Paired strict outcomes for each component removal over the same
200 targets. Each bar partitions targets into cases where both policies were
strictly successful, only PSA-PPO was successful, only the ablated policy was successful, or both failed. $\Delta$ denotes the strict-success rate of
PSA-PPO minus that of the corresponding ablation.}
\label{fig:app_ablation_paired_strict}
\end{figure}

Figure~\ref{fig:app_ablation_paired_strict} exposes the paired strict-outcome
contingencies underlying the aggregate differences in
Table~\ref{tab:app_ablation_paired_binary}.

\begin{figure*}
\centering
\includegraphics[width=0.88\textwidth]{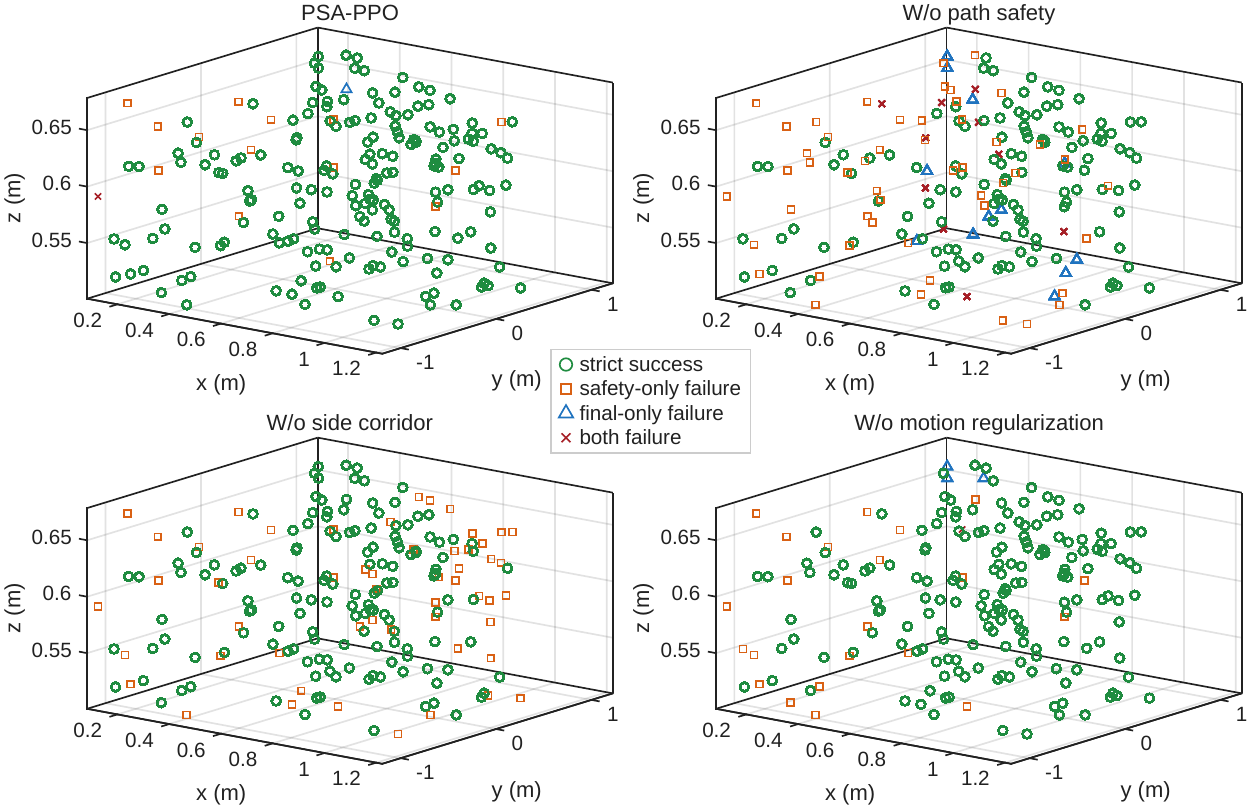}
\caption{Target-wise simulation outcomes over the common workspace for
PSA-PPO and the three component-removal formulations. Markers distinguish
strict success, safety-only failure, final-only failure, and failure of both
criteria. The plots localize evaluation failures but do not represent
variation across policy retraining seeds.}
\label{fig:app_ablation_workspace_outcomes}
\end{figure*}

The three-dimensional outcome maps in
Figure~\ref{fig:app_ablation_workspace_outcomes} retain target location and
failure type simultaneously. Numerical robustness of the selected PSA-PPO
checkpoint across the predefined workspace bins is given in
Table~\ref{tab:app_sim_workspace_numeric}; the corresponding multi-policy
bin comparison is reported in the main paper.

\begin{table*}
\centering
\footnotesize
\caption{Numerical workspace robustness of the selected PSA-PPO checkpoint across predefined equal-width bins.}
\label{tab:app_sim_workspace_numeric}
\setlength{\tabcolsep}{3.5pt}
\renewcommand{\arraystretch}{1.05}
\begin{tabular}{@{}llcccccc@{}}
\toprule
\textbf{Dimension} & \textbf{Interval} & $N$ &
\textbf{Final (\%)} & \textbf{Path-safe (\%)} & \textbf{Strict (\%)} &
\textbf{Strict 95\% CI} &
\multicolumn{1}{c}{\shortstack{\textbf{Median error /}\\\textbf{P95 violation (mm)}}}\\
\midrule
Radius & [0.6000, 0.7625) m & 44 & 97.7 & 95.5 & 93.2 & [81.8, 97.7] & 33.16 / 1.34 \\
Radius & [0.7625, 0.9250) m & 45 & 100.0 & 97.8 & 97.8 & [88.4, 99.6] & 33.63 / 0.00 \\
Radius & [0.9250, 1.0875) m & 47 & 100.0 & 95.7 & 95.7 & [85.8, 98.8] & 29.89 / 0.36 \\
Radius & [1.0875, 1.2500] m & 64 & 98.4 & 84.4 & 84.4 & [73.6, 91.3] & 33.15 / 3.63 \\
\addlinespace
Azimuth & [-85.0, -51.0) deg & 37 & 97.3 & 78.4 & 78.4 & [62.8, 88.6] & 32.40 / 6.03 \\
Azimuth & [-51.0, -17.0) deg & 42 & 100.0 & 95.2 & 95.2 & [84.2, 98.7] & 33.11 / 0.62 \\
Azimuth & [-17.0, 17.0) deg & 39 & 100.0 & 94.9 & 94.9 & [83.1, 98.6] & 31.24 / 1.53 \\
Azimuth & [17.0, 51.0) deg & 43 & 97.7 & 97.7 & 95.3 & [84.5, 98.7] & 31.50 / 0.84 \\
Azimuth & [51.0, 85.0] deg & 39 & 100.0 & 94.9 & 94.9 & [83.1, 98.6] & 34.06 / 0.20 \\
\addlinespace
Height & [0.500, 0.560) m & 64 & 100.0 & 98.4 & 98.4 & [91.7, 99.7] & 33.48 / 0.13 \\
Height & [0.560, 0.620) m & 60 & 98.3 & 91.7 & 91.7 & [81.9, 96.4] & 30.60 / 2.59 \\
Height & [0.620, 0.680] m & 76 & 98.7 & 88.2 & 86.8 & [77.4, 92.7] & 33.09 / 3.40 \\
\bottomrule
\end{tabular}
\end{table*}

\begin{figure*}
\centering
\includegraphics[width=0.92\textwidth]{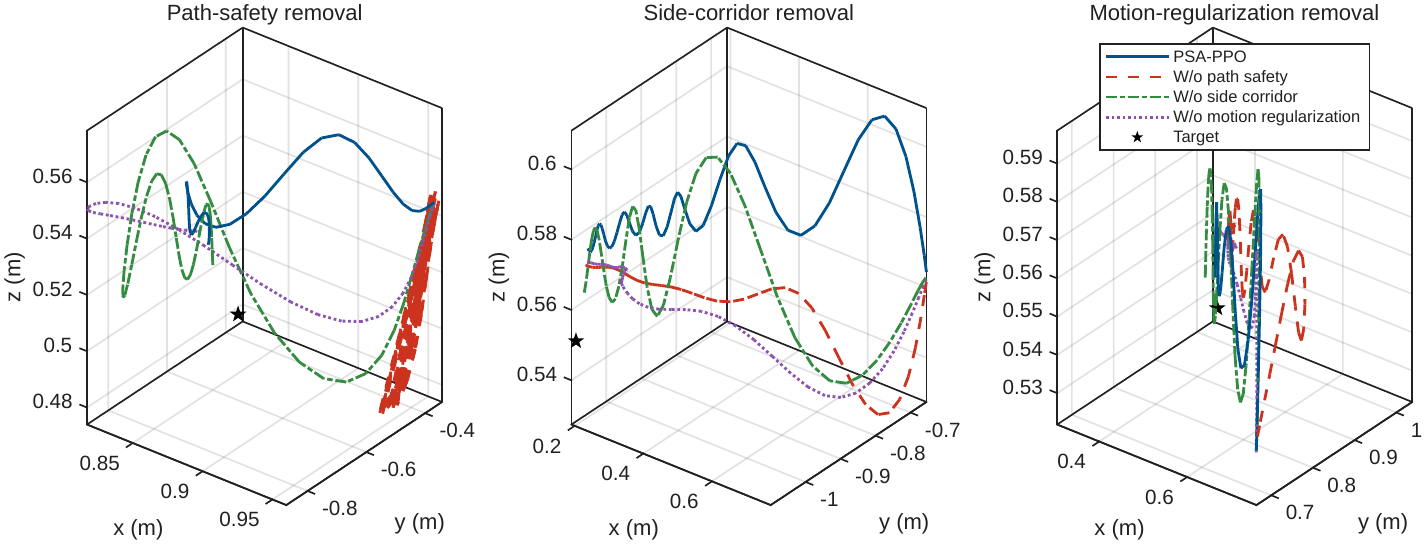}
\caption{Representative trajectory overlays for the three component-removal
contrasts. The displayed targets were selected from the common evaluation set
to expose the corresponding contrast: target 65 for path-safety removal,
target 31 for side-corridor removal, and target 199 for motion-regularization
removal. The star marks the target location. These examples supplement, rather
than replace, the paired population statistics.}
\label{fig:app_ablation_representative}
\end{figure*}

Representative target-matched trajectories selected for each component
contrast are shown in Figure~\ref{fig:app_ablation_representative}. They are
included as qualitative diagnostics and are not used to estimate population
effects.

\begin{figure*}
\centering
\includegraphics[width=0.92\textwidth]{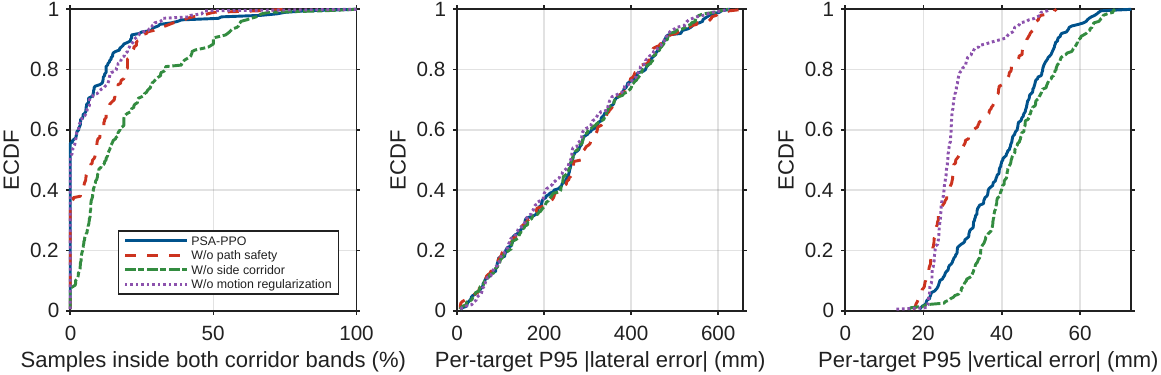}
\caption{Post-hoc fixed-band corridor diagnostics over the common target set:
fraction of trajectory samples inside both lateral and vertical bands,
per-target P95 absolute lateral error, and per-target P95 absolute vertical
error. The distributions do not exhibit a uniform ordering across
formulations and should not be interpreted as a direct measurement of the
complete corridor reward.}
\label{fig:app_ablation_corridor}
\end{figure*}

The fixed-band corridor measures in
Figure~\ref{fig:app_ablation_corridor} were reconstructed post hoc from the
recorded trajectories. They are not mathematically identical to
$\mathcal{R}^{cor}_t$, which also contains continuous progress, overshoot,
staging, and target-coupled terms. They are therefore reported as secondary
diagnostics rather than as an isolated decomposition of the corridor reward.

\begin{figure*}
\centering
\includegraphics[width=0.92\textwidth]{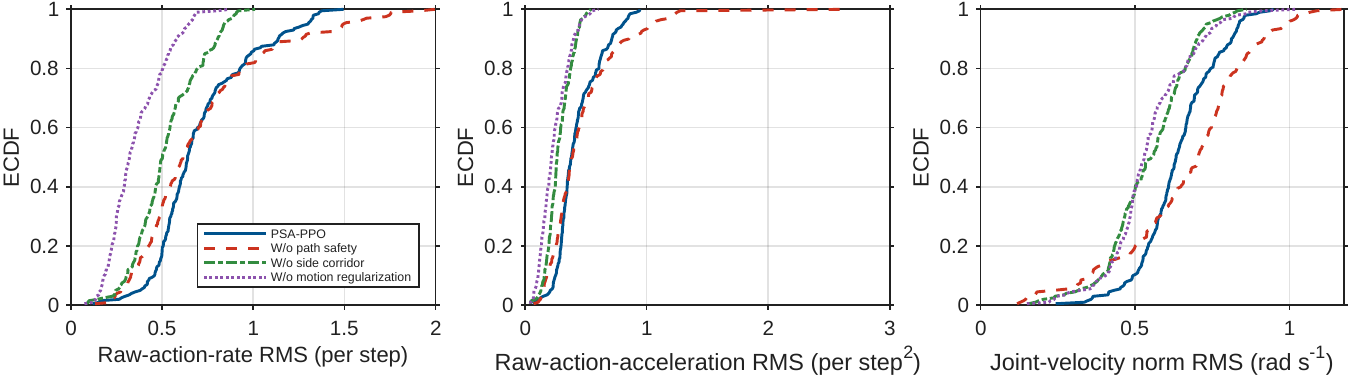}
\caption{Empirical distributions of direct motion-regularization diagnostics:
raw-action-rate RMS, raw-action-acceleration RMS, and joint-velocity-norm RMS.
The no-motion-regularization policy does not shift these quantities uniformly
toward larger values, demonstrating that the component removal changes the
learned operating point rather than providing a monotonic smoothness test.}
\label{fig:app_ablation_motion_reg}
\end{figure*}

Figure~\ref{fig:app_ablation_motion_reg} reports quantities aligned directly
with the removed motion-regularization terms. The no-motion-regularization
checkpoint does not produce a monotonic increase in these quantities, which is
why the main paper does not attribute lower TCP jerk causally to
$\mathcal{R}^{reg}_t$.

\begin{figure*}
\centering
\includegraphics[width=0.82\textwidth]{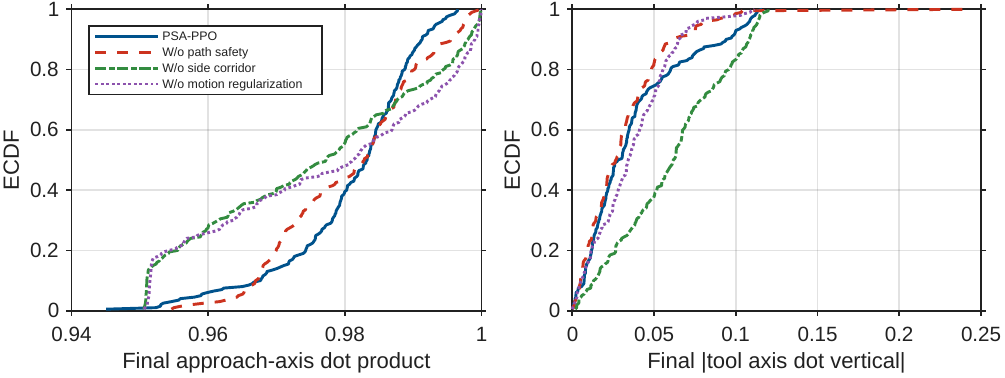}
\caption{Final tool-orientation distributions over the common evaluation set:
approach-axis alignment (left) and absolute tool-axis vertical component
(right). Orientation terms and final orientation criteria were unchanged
across the three component-removal formulations.}
\label{fig:app_ablation_orientation}
\end{figure*}

Tool-orientation behavior is provided in
Figure~\ref{fig:app_ablation_orientation}. Orientation shaping and the final
orientation criteria were retained in all three ablations.

\FloatBarrier
\section{Physical Controller Benchmark and Deployment Details}
\label{app:C}

This appendix documents the controller configurations, measured-trajectory
processing, analysis populations, and secondary diagnostics supporting the
physical comparison in Section~\ref{subsubsec:inhouse_controller_comparison}.
Binary reaching outcomes use all scheduled targets for each controller,
whereas continuous motion comparisons use the common-success target population
defined below.

\subsection{Physical Execution and Planner Configuration}
\label{app:C_physical_config}

The controlled benchmark used the same physical target definitions for
PSA-PPO, Cartesian+TOTG, RRTConnect, and RRT*. The paper-level reaching
criterion was a 20-mm Cartesian tolerance. The online controller monitor used
a tighter 2-mm threshold for execution management; this monitor threshold was
not used to define the reported task success.

\begin{table*}
\centering
\footnotesize
\caption{Principal physical execution and logging parameters used for
PSA-PPO deployment and the controlled real-robot benchmark.}
\label{tab:app_physical_execution}
\setlength{\tabcolsep}{4.5pt}
\renewcommand{\arraystretch}{1.08}
\begin{tabular}{@{}lll@{}}
\toprule
\textbf{Parameter} & \textbf{Value} & \textbf{Role} \\
\midrule
HOME TCP position &
$[-0.164,\;0.641,\;0.382]$ m &
Robot observation/home pose \\

Benchmark tool orientation $(w,x,y,z)$ &
$[0.009162454,\;0.005118748,\;0.718754832,\;0.695184393]$ &
Common fixed physical orientation \\

Base frame &
\texttt{robot\_arm\_base\_link} &
Cartesian reference frame \\

Tool frame &
\texttt{robot\_arm\_tool0} &
Measured TCP frame \\

Paper evaluation tolerance &
0.020 m &
Binary target-reaching criterion \\

ROS monitor tolerance &
0.002 m &
Online execution-monitor threshold \\

Consecutive monitor checks &
5 &
Required in-tolerance monitor observations \\

TCP/TF sampling rate &
100 Hz &
Measured trajectory sampling \\

Endpoint window &
0.50 s &
Terminal TCP estimation window \\

Motion timeout &
12.0 s &
Maximum PSA-PPO reach command interval \\

Acceleration pre-check threshold &
3.0 rad/s$^2$ &
Pre-command hardware safeguard \\

Acceleration trip threshold &
3.0 rad/s$^2$ &
Feedback-based HOLD/HALT threshold \\

Time-shaping acceleration &
0.0045 rad/s$^2$ &
PSA-PPO trajectory-duration shaping limit \\
\bottomrule
\end{tabular}
\end{table*}

PSA-PPO retained the frozen deployment correction described in
Section~\ref{subsec:ros_harvesting_control}. Table~\ref{tab:app_physical_execution} summarizes the execution and logging
parameters used for PSA-PPO deployment on the UR10e. Planner targets were supplied directly to MoveIt without that policy-specific Cartesian correction. Accordingly, the benchmark compares the complete deployed controllers under common target definitions rather than claiming identical low-level command
generation.

The MoveIt baselines used a common staged physical benchmark harness. For each
stage, the measured current joint state was supplied as the planner start state
when available, and a planned-start mismatch above 0.005~rad prevented
execution. OMPL-generated trajectories were collision checked by MoveIt and
then sent directly to the UR scaled joint-trajectory controller. Cartesian
motion used collision-checked Cartesian interpolation followed by
time-optimal trajectory generation (TOTG). The principal planner settings are
listed in Table~\ref{tab:app_planner_settings}.

\begin{table*}
\centering
\footnotesize
\caption{MoveIt baseline configuration used in the controlled physical
benchmark. Parameters listed as common were applied to the planner-based
controllers unless the backend required a method-specific setting.}
\label{tab:app_planner_settings}
\setlength{\tabcolsep}{4.2pt}
\renewcommand{\arraystretch}{1.08}
\begin{tabular}{@{}p{3.5cm}p{3.8cm}p{8.1cm}@{}}
\toprule
\textbf{Setting} & \textbf{Value} & \textbf{Application} \\
\midrule

Move group &
\texttt{arm} &
All MoveIt baselines \\

RRTConnect backend &
OMPL RRTConnect (\texttt{RRTConnectkConfigDefault}) &
Pose planning using the frozen RRTConnect planner ID \\

RRT* backend &
OMPL RRT* (\texttt{RRTstar}) &
Pose planning using the post-fix RRT* planner ID \\

Cartesian backend &
Cartesian linear + TOTG (\texttt{CartesianLinearTOTG}) &
Collision-checked Cartesian interpolation followed by retiming \\

Planning time &
8.0 s &
OMPL pose planners \\

Planning attempts &
3 &
OMPL pose planners \\

Velocity scaling &
0.10 &
MoveIt trajectory generation/retiming \\

Acceleration scaling &
0.10 &
MoveIt trajectory generation/retiming \\

Goal-position tolerance &
0.002 m &
MoveIt pose target \\

Goal-orientation tolerance &
0.035 rad &
MoveIt pose target \\

Maximum start-state mismatch &
0.005 rad &
Measured-versus-planned joint start check \\

Cartesian EEF step &
0.005 m &
Cartesian interpolation \\

Minimum Cartesian fraction &
0.999 &
Required fraction for Cartesian plan acceptance \\

Cartesian collision checking &
Enabled &
\texttt{avoid\_collisions=True} \\

Cartesian retiming &
TOTG &
\texttt{time\_optimal\_trajectory\_generation} \\

Replanning &
Disabled &
All MoveIt baselines \\

Planner safe-height offset &
0.035 m &
$z_{\mathrm{safe}}=z^T+0.035$ m for the staged planner approach \\

Planner far/near backoff &
$-0.040/-0.020$ m &
Intermediate staged approach offsets along the benchmark $y$ direction \\

Execution interface &
\texttt{FollowJointTrajectory} &
\texttt{/robot/arm/scaled\_pos\_traj\_controller/}
\texttt{follow\_joint\_trajectory} \\
\bottomrule
\end{tabular}
\end{table*}

The final RRT* batch wrapper explicitly passed \texttt{planner\_id = RRTstar} and verified that the corresponding ROS planner
configuration resolved to an RRT* implementation before execution. The
earlier \texttt{RRTstarkConfigDefault} wrapper was superseded and is not used
as the provenance for the reported RRT* configuration.

\subsection{Trajectory Reconstruction and Analysis Populations}
\label{app:C_processing}

TCP transforms and joint/controller states were archived at 100~Hz for all
controllers. All cross-controller kinematic and trajectory-geometry quantities
reported in the paper were recomputed offline from these raw trajectory
archives using one common processing pipeline; the online per-segment
path/speed/acceleration/jerk estimates produced during data collection were
not used for the comparative analysis because invalid online path-length
estimates were observed for some RRTConnect segments. In the common offline
pipeline, samples were timestamp ordered, resampled to a uniform grid, and
smoothed using a 0.15-s moving-average window before derivative-based
analysis. Trajectory length was evaluated from the smoothed position stream at
30~Hz, with a 2-mm Ramer--Douglas--Peucker tolerance used for geometric
simplification. TCP velocity, acceleration, and jerk were obtained by central
finite differences. The reported high-end jerk statistic is the 99.5th
percentile rather than the numerical maximum. The final 0.22-m
target-relative window was fixed from the local PSA-PPO task definition before
controller comparison and was applied identically to all methods.

Binary reliability and continuous trajectory comparisons use different
populations by construction. All scheduled targets contribute to binary
reaching outcomes. Continuous controller comparisons condition on targets
retained as successful for all four controllers, so that each paired metric
compares the same physical target configuration. Table~\ref{tab:app_physical_populations}
summarizes these populations.

\begin{table*}
\centering
\footnotesize
\caption{Physical benchmark populations used for binary and continuous
controller comparisons.}
\label{tab:app_physical_populations}
\setlength{\tabcolsep}{5.5pt}
\renewcommand{\arraystretch}{1.08}
\begin{tabular}{@{}lcccc@{}}
\toprule
&
\textbf{PSA-PPO} &
\textbf{Cartesian+TOTG} &
\textbf{RRTConnect} &
\textbf{RRT*} \\
\midrule
Scheduled targets & 150 & 150 & 150 & 150 \\
Archived measured trajectories & 150 & 150 & 150 & 125 \\
Paper-level reached targets & 142 & 129 & 132 & 120 \\
Paper-level failed targets & 8 & 21 & 18 & 30 \\
\bottomrule
\end{tabular}

\vspace{2pt}
\begin{minipage}{0.97\textwidth}
\footnotesize
The 25 RRT* targets without measured trajectory archives remain failures in
the binary denominator and are missing, rather than imputed, for continuous
metrics. The four-controller common-success population used for the primary
continuous analyses contains $N=97$ targets after the witnessed-count
sensitivity protocol described in Section~\ref{app:C_sensitivity}.
\end{minipage}
\end{table*}

The physical target was treated as the experimental unit. Paired continuous
differences were evaluated using Wilcoxon signed-rank tests and 10,000 paired
bootstrap resamples for confidence intervals. Holm correction was applied
within each prespecified family of PSA-PPO--planner comparisons. Trajectory
samples within a target were not treated as independent observations.

\subsection{Detailed Motion and Geometry Statistics}
\label{app:C_motion_details}

The main paper emphasizes the final 0.22-m local region because this matches
the learned crop-proximal control scope. Table~\ref{tab:app_full_approach_metrics}
shows that the same qualitative dynamic ordering is also present over the
complete reconstructed approach.

\begin{table*}
\centering
\footnotesize
\caption{Selected physical motion metrics over the complete reconstructed
approach on the common-success target set ($N=97$). Values are controller
medians. Lower values indicate lower dynamic excitation or tracking error.}
\label{tab:app_full_approach_metrics}
\setlength{\tabcolsep}{7pt}
\renewcommand{\arraystretch}{1.10}
\begin{tabular}{@{}lcccc@{}}
\toprule
\textbf{Metric} &
\textbf{PSA-PPO} &
\textbf{Cartesian+TOTG} &
\textbf{RRTConnect} &
\textbf{RRT*} \\
\midrule
RMS acceleration (m/s$^2$) &
0.1776 & 0.5776 & 0.5856 & 0.2739 \\

RMS jerk (m/s$^3$) &
15.82 & 47.68 & 45.01 & 19.91 \\

99.5th-percentile jerk (m/s$^3$) &
105.11 & 143.38 & 172.29 & 127.58 \\

Joint-position tracking RMSE (rad) &
$8.41\times10^{-4}$ &
$2.87\times10^{-3}$ &
$2.76\times10^{-3}$ &
$1.31\times10^{-3}$ \\
\bottomrule
\end{tabular}
\end{table*}

Target-wise effect sizes for the final 0.22-m region are reported in Table~\ref{tab:app_local_paired_effects}. Differences are defined as PSA-PPO minus planner, so negative values indicate a lower PSA-PPO metric. The endpoint-error ECDF is provided in the main paper; the numerical medians are retained in Table~\ref{tab:app_endpoint_accuracy} for direct lookup. Population-level trajectory geometry is reported in Table~\ref{tab:app_trajectory_geometry} because lower dynamic excitation does not imply a shorter or straighter path.

\begin{table*}
\centering
\footnotesize
\caption{Paired effect-size analysis for PSA-PPO relative to each classical
planner within the final 0.22-m target-relative approach. Differences are
computed target-wise over the identical common-success set ($N=97$).}
\label{tab:app_local_paired_effects}
\setlength{\tabcolsep}{4.0pt}
\renewcommand{\arraystretch}{1.08}
\begin{tabular}{@{}llccc@{}}
\toprule
\textbf{Metric} &
\textbf{Planner} &
\textbf{PSA-PPO reduction (\%)} &
\textbf{Median paired difference [95\% CI]} &
\textbf{Holm-adjusted $p$} \\
\midrule

RMS acceleration (m/s$^2$) &
Cartesian+TOTG & 75.9 &
$-0.4012$ [$-0.4118,-0.3953$] &
$4.02\times10^{-16}$ \\
&
RRTConnect & 77.1 &
$-0.4295$ [$-0.4419,-0.4162$] &
$4.02\times10^{-16}$ \\
&
RRT* & 52.4 &
$-0.1332$ [$-0.1398,-0.1303$] &
$4.02\times10^{-16}$ \\
\addlinespace

RMS jerk (m/s$^3$) &
Cartesian+TOTG & 73.0 &
$-30.71$ [$-31.79,-29.83$] &
$4.02\times10^{-16}$ \\
&
RRTConnect & 72.1 &
$-29.34$ [$-30.62,-28.54$] &
$4.02\times10^{-16}$ \\
&
RRT* & 39.8 &
$-7.177$ [$-7.625,-6.442$] &
$4.02\times10^{-16}$ \\
\addlinespace

Jerk $P_{99.5}$ (m/s$^3$) &
Cartesian+TOTG & 41.1 &
$-57.80$ [$-63.84,-53.18$] &
$4.02\times10^{-16}$ \\
&
RRTConnect & 52.8 &
$-91.01$ [$-98.11,-80.84$] &
$4.02\times10^{-16}$ \\
&
RRT* & 39.2 &
$-49.96$ [$-55.02,-44.42$] &
$4.02\times10^{-16}$ \\
\addlinespace

Joint tracking RMSE (rad) &
Cartesian+TOTG & 72.8 &
$-1.840\times10^{-3}$
[$-1.868,-1.804$]$\times10^{-3}$ &
$4.02\times10^{-16}$ \\
&
RRTConnect & 71.8 &
$-1.785\times10^{-3}$
[$-1.816,-1.742$]$\times10^{-3}$ &
$4.02\times10^{-16}$ \\
&
RRT* & 39.4 &
$-4.629\times10^{-4}$
[$-4.912,-4.179$]$\times10^{-4}$ &
$4.02\times10^{-16}$ \\
\bottomrule
\end{tabular}

\vspace{2pt}
\begin{minipage}{0.98\textwidth}
\footnotesize
Reduction percentages are calculated from the controller medians. Confidence
intervals are paired target-level bootstrap intervals. Tests are paired
Wilcoxon signed-rank tests with Holm correction within the prespecified
local-motion comparison family.
\end{minipage}
\end{table*}

\begin{table}
\centering
\footnotesize
\caption{Terminal positioning accuracy on the common-success physical target
set. The physical reaching tolerance was 20~mm.}
\label{tab:app_endpoint_accuracy}
\setlength{\tabcolsep}{5.0pt}
\renewcommand{\arraystretch}{1.08}
\begin{tabular}{@{}lcc@{}}
\toprule
\textbf{Controller} &
\shortstack{\textbf{Median TCP}\\\textbf{error (mm)}} &
\shortstack{\textbf{Fraction of 20-mm}\\\textbf{tolerance (\%)}} \\
\midrule
PSA-PPO        & 9.78 & 48.9 \\
Cartesian+TOTG & 0.06 & 0.3 \\
RRTConnect     & 1.62 & 8.1 \\
RRT*           & 1.61 & 8.1 \\
\bottomrule
\end{tabular}
\end{table}

\begin{table*}[t]
\centering
\footnotesize
\caption{Trajectory-geometry and spatial-excursion statistics over the full
reconstructed physical approach. Median quantities characterize typical path
geometry; $P_{95}$ and maximum deviations characterize distribution tails.}
\label{tab:app_trajectory_geometry}
\setlength{\tabcolsep}{4.2pt}
\renewcommand{\arraystretch}{1.08}
\begin{tabular}{@{}lcccc@{}}
\toprule
\textbf{Metric} &
\textbf{PSA-PPO} &
\textbf{Cartesian+TOTG} &
\textbf{RRTConnect} &
\textbf{RRT*} \\
\midrule
Detour ratio & 1.8094 & 1.6697 & 1.6812 & 1.6962 \\
Path efficiency & 0.5527 & 0.5989 & 0.5948 & 0.5896 \\
Backtracking ratio & 0.0371 & 0.0178 & 0.0195 & 0.0218 \\
Total Cartesian turning ($^\circ$) & 410.09 & 16.91 & 75.58 & 75.14 \\
Base-azimuth travel ($^\circ$) & 26.25 & 25.73 & 25.97 & 26.45 \\
Total joint travel (rad) & 3.108 & 2.850 & 2.910 & 2.723 \\
$P_{95}$ maximum line deviation (mm) & 235 & 235 & 237 & 235 \\
Maximum line deviation (mm) & 292 & 297 & 297 & 297 \\
$P_{95}$ base-azimuth travel ($^\circ$) & 61.1 & 61.2 & 61.5 & 61.6 \\
Maximum base-azimuth travel ($^\circ$) & 66.9 & 66.8 & 67.1 & 67.0 \\
Full-base-revolution events & 0/97 & 0/97 & 0/97 & 0/97 \\
Any full-joint-revolution events & 0/97 & 0/97 & 0/97 & 0/97 \\
\bottomrule
\end{tabular}

\vspace{2pt}
\begin{minipage}{0.98\textwidth}
\footnotesize
Median statistics use the common-success target set ($N=97$). For detour
ratio, backtracking, turning, azimuth travel, joint travel, and line deviation,
lower values indicate more compact motion; for path efficiency, higher values
indicate greater geometric directness. Tail statistics are descriptive and
do not constitute collision or safety-event labels.
\end{minipage}
\end{table*}

\subsection{Failure-Identity Sensitivity and Timing Semantics}
\label{app:C_sensitivity}

The aggregate physical reaching totals were witnessed during the experiment,
but the exact target identities of some additional planner failures were not
independently preserved. These counts were therefore not retrospectively
treated as recovered target-level ground truth. Instead, a deterministic
sensitivity analysis retained the witnessed aggregate totals and selected the
additional planner failure identities from originally logged-success trials
using telemetry anomaly rank.

Hard diagnostic events were prioritized when tool-orientation travel reached
at least $360^\circ$, any single-joint travel reached at least $2\pi$, or the
wrist/elbow angular margin to the nearest multiple of $\pi$ was at most
$5^\circ$. Remaining trials were ranked within each planner using percentile
tails of trajectory geometry, orientation travel, joint excursion,
acceleration, jerk, tracking error, and elapsed duration. The composite score
was
\begin{equation}
S_k
=
0.60\,\bar{P}_k
+
0.25\,P_k^{\max}
+
0.15\,I_k^{\mathrm{hard}},
\label{eq:app_anomaly_score}
\end{equation}
\noindent where $\bar{P}_k$ and $P_k^{\max}$ denote the mean and maximum
within-planner diagnostic percentiles, respectively. Target ID was used to
break ties. The selected identities were used only to form a sensitivity
population consistent with the witnessed counts; they are not claimed as independently observed failure identities. Table~\ref{tab:app_sensitivity_summary} summarizes the witnessed failure counts, trajectory availability, and anomaly-ranked identity selections used
to construct this sensitivity population.

\begin{table*}
\centering
\footnotesize
\caption{Summary of the anomaly-ranked witnessed-count sensitivity
construction for the planner baselines.}
\label{tab:app_sensitivity_summary}
\setlength{\tabcolsep}{5pt}
\renewcommand{\arraystretch}{1.08}
\begin{tabular}{@{}lccccl@{}}
\toprule
\textbf{Planner} &
\shortstack{\textbf{Witnessed}\\\textbf{failures}} &
\shortstack{\textbf{Additional ranked}\\\textbf{identities selected}} &
\shortstack{\textbf{Archived}\\\textbf{trajectories}} &
\shortstack{\textbf{Hard-event}\\\textbf{selected trials}} &
\shortstack{\textbf{Highest-priorityt}\\\textbf{diagnostic target}}\\
\midrule
Cartesian+TOTG & 21 & 19 & 150 & 0 & T119 \\
RRTConnect     & 18 & 18 & 150 & 3 & T040 \\
RRT*           & 30 & 5  & 125 & 4 & T087 \\
\bottomrule
\end{tabular}

\vspace{2pt}
\begin{minipage}{0.98\textwidth}
\footnotesize
The 25 unarchived RRT* attempts remain failures for binary outcome accounting
and are unavailable for continuous trajectory metrics. ``Hard-event selected
trials'' denotes selected cases containing at least one of the predefined
orientation-revolution, single-joint-revolution, or near-singularity
diagnostics. The resulting corrected four-controller common-success population
contains 97 targets.
\end{minipage}
\end{table*}

Timing variables from the benchmark are retained only as harness diagnostics.
The PSA-PPO harness issued multiple external commands and monitored the TCP
against the 2-mm execution threshold, whereas the paper-level outcome uses
20~mm. Consequently, elapsed benchmark time includes monitoring and waiting
behavior and cannot be interpreted as neural-network inference latency or as
intrinsic controller execution speed. The audit found that only 32 of 1080
PSA-PPO harness commands reached the 2-mm monitor threshold; 1048 did not
reach that threshold before monitor completion or timeout, and all 142
paper-level successful PSA-PPO targets contained at least one such inferred
2-mm timeout. For this reason, timing is classified as
\emph{benchmark-harness-dependent} and is excluded from controller-speed
claims.

\subsection{Kinematic and Atypical-Trajectory Diagnostics}
\label{app:C_tail_diagnostics}

\begin{figure}
\centering
\includegraphics[width=0.95\linewidth]
{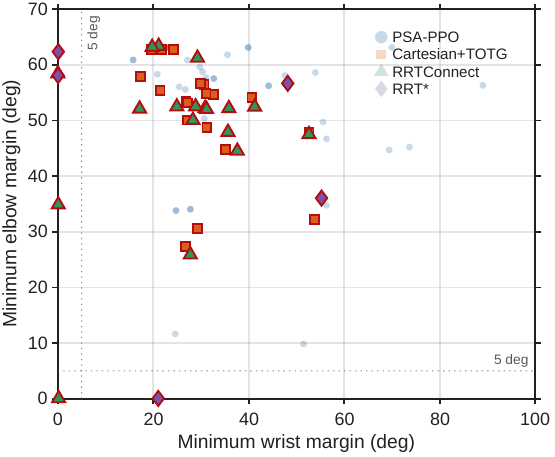}
\caption{Minimum wrist- and elbow-angle margins to the nearest multiple of
$\pi$ in the physical benchmark. The dashed $5^\circ$ reference denotes the
diagnostic threshold used by the anomaly-ranking sensitivity analysis. The
quantity is kinematic evidence only; near-threshold configurations are not
interpreted as proof of singularity-induced failure.}
\label{fig:app_singularity_margin}
\end{figure}

Singularity proximity was evaluated as a diagnostic rather than a causal
failure label. Wrist and elbow margins were defined as the minimum absolute
angular distance to a multiple of $\pi$ for UR joint~5 and joint~3,
respectively. The $5^\circ$ threshold in
Figure~\ref{fig:app_singularity_margin} was used only by the anomaly-ranking
audit; proximity to this threshold does not demonstrate that a singularity
caused a stall or failed reach.

Within the sensitivity-selected sets, Cartesian+TOTG contained no
orientation-revolution, single-joint-revolution, or near-singularity hard
events. The RRTConnect selected set contained two orientation-revolution
events, one single-joint-revolution event, three wrist-near-singularity cases,
and one elbow-near-singularity case. The RRT* selected set contained four,
three, two, and one such cases, respectively. These counts describe the
selected diagnostic subsets and not the $N=97$ population as a whole.

\begin{figure*}
\centering
\subfloat[]{
    \includegraphics[width=0.315\textwidth]
    {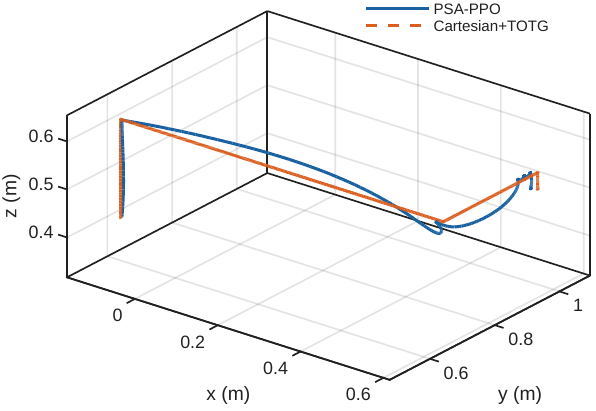}
    \label{fig:app_unusual_cartesian}}
\hfill
\subfloat[]{
    \includegraphics[width=0.315\textwidth]
    {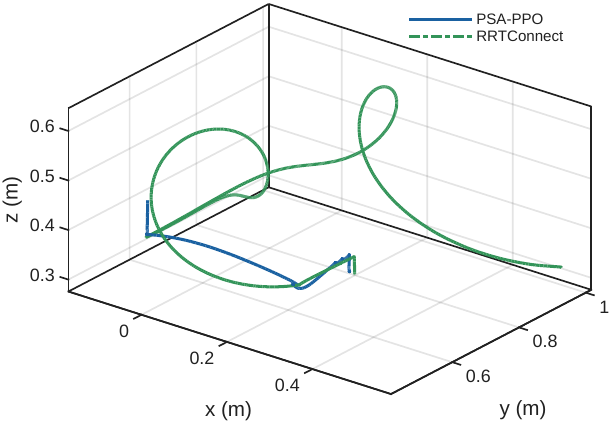}
    \label{fig:app_unusual_rrtconnect}}
\hfill
\subfloat[]{
    \includegraphics[width=0.315\textwidth]
    {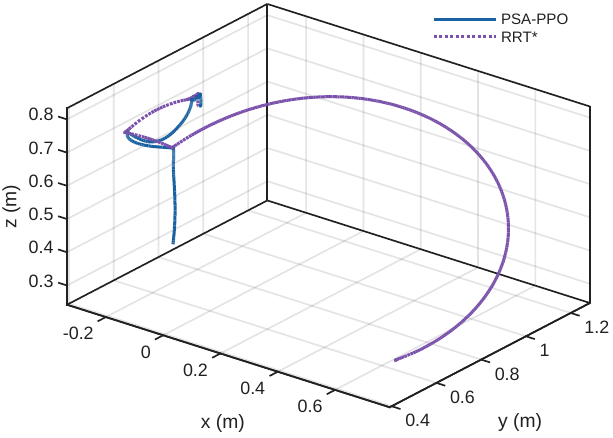}
    \label{fig:app_unusual_rrtstar}}
\caption{Matched target-level trajectory diagnostics from the
anomaly-ranked sensitivity analysis: (a) PSA-PPO and Cartesian+TOTG on T119,
(b) PSA-PPO and RRTConnect on T040, and (c) PSA-PPO and RRT* on T087.
The examples are the highest-priority selected planner cases for their
respective methods and are not representative population trajectories.
Spatially expansive motion and near-singularity diagnostics are not
interpreted as collision events or as proof of failure causation.}
\label{fig:app_matched_unusual}
\end{figure*}

Figure~\ref{fig:app_matched_unusual} shows the highest-priority selected
planner execution together with PSA-PPO on the same target. The Cartesian
case is the highest-ranked non-hard anomaly because no Cartesian hard event
was present in the selected set. The RRTConnect and RRT* examples contain
hard diagnostic events under the predefined audit rule. These examples
illustrate tail behavior only and are not classified as collision events.

\end{document}